\documentclass[sn-mathphys]{sn-jnl}

\jyear{2022}%

\theoremstyle{thmstyleone}%
\newtheorem{proposition}{Proposition}%
\newtheorem{proof1}{Proof}%
\newtheorem{lemma}{Lemma}%

\theoremstyle{thmstyletwo}%
\newtheorem{example}{Example}%

\theoremstyle{thmstylethree}%
\newtheorem{definition}{Definition}%

\definecolor{green}{rgb}{0.0, 0.5, 0.0}

\newcommand{\name}{\textsc{MagicPopper}}
\newcommand{\popper}{\textsc{Popper}}
\newcommand{\metagol}{\textsc{Metagol}}

\newcommand{\ale}{\textsc{Aleph}}
\newcommand{\til}{\textsc{Tilde}}
\newcommand{\progol}{\textsc{Progol}}

\newcommand{\deltailp}{\textsc{$\delta$-ILP}}
\newcommand{\aspal}{\textsc{Aspal}}
\newcommand{\popperplus}{\textsc{Popper+}}

\usepackage{pgfkeys}
    \newenvironment{customlegend}[1][]{%
        \begingroup
        \csname pgfplots@init@cleared@structures\endcsname
        \pgfplotsset{#1}%
    }{%
        \csname pgfplots@createlegend\endcsname
        \endgroup
    }%
    \def\addlegendimage{\csname pgfplots@addlegendimage\endcsname}
\usepackage{comment}

\usepackage{booktabs}
\usepackage{inconsolata}

\usepackage{textcomp}
\usepackage{amsmath,amsthm}
\usepackage[cmintegrals]{newtxsf}

\begin{document}

\title{Learning programs with magic values}


\author*[1]{\fnm{Céline} \sur{Hocquette}}\email{celine.hocquette@cs.ox.ac.uk}

\author[1]{\fnm{Andrew} \sur{Cropper}}\email{andrew.cropper@cs.ox.ac.uk}


\affil[1]{\orgdiv{Department of Computer Science}, \orgname{University of Oxford}}





\abstract{A magic value in a program is a constant symbol that is essential for the execution of the program but has no clear explanation for its choice. Learning programs with magic values is difficult for existing program synthesis approaches. To overcome this limitation, we introduce an inductive logic programming approach to efficiently learn programs with magic values. Our experiments on diverse domains, including program synthesis, drug design, and game playing, show that our approach can (i) outperform existing approaches in terms of predictive accuracies and learning times, (ii) learn magic values from infinite domains, such as the value of pi, and (iii) scale to domains with millions of constant symbols.}




\maketitle

\section{Introduction}

A magic value in a program is a constant symbol that is essential for the good execution of the program but has no clear explanation for its choice. 
For instance, consider the problem of classifying lists.
Figure \ref{fig:intro_ex} shows positive and negative examples.
Figure \ref{fig:intro_target} shows a hypothesis which  discriminates between the positive and negative examples.
Learning this hypothesis involves the identification of the magic number \emph{7}.

\begin{figure}[ht]
\centering
\begin{tabular}{ll}
Positive examples & Negative examples\\
\midrule
\texttt{f([a,b,c,7,8,k,f])} & \texttt{f([x,y,z,8,3,k,f,x,t,8,k,f])} \\
\texttt{f([p,3,9,y,5,r,a,q,7])} &  \texttt{f([a,b,c,p,r,w,q,9])}\\
\texttt{f([e,a,a,b,c,7,t,a,b,c,x,z,r,t])} & \texttt{f([u,k,a,b,c,z,r,t,5,e,t])}
\end{tabular} 
\caption{Positive and negative examples}\label{fig:intro_ex}

\begin{tabular}{l}
\emph{f(A)$\leftarrow$ head(A,\textbf{7})}\\
\emph{f(A)$\leftarrow$ tail(A,B),f(B)}\\
\end{tabular}
    \caption{Target hypothesis}
    \label{fig:intro_target}

\begin{tabular}{l}
\emph{f(A)$\leftarrow$ head(A,B), @magic(B)}\\
\emph{f(A)$\leftarrow$ tail(A,B), f(B)}\\
\end{tabular}
    \caption{Intermediate hypothesis}
    \label{fig:intro_intermediate}
\end{figure}

Magic values are fundamental to many areas of knowledge, including physics and mathematics. For instance, the value of \emph{pi} is essential to compute the area of a disk.
Likewise, the gravitational constant is essential to  identify whether an object subject to its weight is in mechanical equilibrium.
Similarly, consider the classical AI task of learning to play games. To play the game \emph{connect four}\footnote{Connect four is a two-player game in which the players take turns dropping coloured tokens into a grid. The goal of the game is to be the first to form a horizontal, vertical, or diagonal line of four of one's own tokens.}, a learner must correctly understand the rules of this game, which implies that they must discover the magic value \emph{four}, i.e. four tokens in a row.

Although fundamental to AI, learning programs with magic values is difficult for existing program synthesis approaches.
For instance, many recent inductive logic programming (ILP) \cite{muggleton1991,cropper2020c} approaches first enumerate all possible rules allowed in a program \cite{aspal,kaminski2018,raghothaman2019,evans2018} and then search for a subset of them.
For example, ASPAL \cite{aspal} precomputes every possible rule and uses an answer set solver to find a subset of them. 
Other approaches similarly represent constants as unary predicate symbols \cite{evans2018,cropper2020b}.
Both approaches suffer from two major limitations.
First, they need a finite and tractable number of constant symbols to search through, which is clearly infeasible for large and infinite domains, such as when reasoning about continuous values.
Second, they might generate rules with irrelevant magic values that never appear in the data, and thus suffer from performance issues.
Older ILP approaches similarly struggle with magic values. 
For instance, for \progol{} \cite{muggleton1995} to learn a rule with a constant symbol, that constant must appear in the bottom clause of an example.
\progol{}, therefore, struggles to learn recursive programs with constant values. 
It can also struggle when the bottom clause grows extremely large due to many potential magic values. 




The goal of this paper, and therefore its main contribution, is to overcome these limitations by introducing an ILP approach that can efficiently learn programs with magic values, including values from infinite and continuous domains.
The key idea of our approach, which is heavily inspired by \ale{}'s lazy evaluation approach  \cite{srinivasan1999}, is to not enumerate all possible magic values but to instead generate hypotheses with variables in place of constant symbols that are later filled in by a learner.
In other words, the learner first builds a partial general hypothesis and then lazily fills in the specific details (the magic values) by examining the given data. For instance, reconsider the task of identifying the magic number \emph{7} in a list. The learner first constructs a partial intermediate hypothesis as the one shown in Figure \ref{fig:intro_intermediate}. In the first clause, the first-order variable $B$ is marked as a constant with the internal predicate \emph{@magic}. However, it is not bound to any particular constant symbol. The value for this magic variable is lazily identified by executing this hypothesis on the examples.

As the example in Figure \ref{fig:intro_intermediate} illustrates, the key advantages of our approach compared to existing ones are that it (i) does not rely on enumeration of all constant symbols but only considers candidate constant values which can be obtained from the examples, (ii) can learn programs with magic values from large and infinite domains, and (iii) can learn magic values for recursive programs.

To implement our approach, we build on the \emph{learning from failures} (LFF) \cite{cropper2020b} approach.
LFF is a constraint-driven ILP approach where the goal is to accumulate constraints on the hypothesis space.
A LFF learner continually generates and tests hypotheses, from which it infers constraints. For instance, if a hypothesis is too general (i.e. entails a negative example), then a generalisation constraint prunes generalisations of this hypothesis from the hypothesis space.

Current LFF approaches \cite{cropper2021,cropper2021b,hopper} cannot, however, reason about partial hypotheses, such as the one shown in Figure \ref{fig:intro_intermediate}. 
They must instead all enumerate constant symbols by representing them as unary predicate symbols.
Current approaches, therefore, suffer from the same limitations as other recent ILP approaches, i.e. they struggle to scale to large and infinite domains.
We, therefore, extend the LFF constraints to prune such intermediate partial hypotheses. 
Each constraint prunes sets of intermediate hypotheses, each of which represents the set of its instantiations. 
We prove that these extended constraints are optimally sound: they do not prune optimal solutions from the hypothesis space. 

We implement our magic value approach in \name, which, as it builds on the LFF learner \popper{}, supports predicate invention \cite{cropper2021} and learning recursive programs.
\name{} can learn programs with magic values from domains with millions of constant symbols and scale to infinite domains.
For instance, we show that \name{} can learn (an approximation of) the value of pi.
In addition, in contrast to existing approaches, \name{} does not need to be told which arguments may be bound to magic values but instead can automatically identify them if any is needed, although this fully automatic approach comes with a high cost in terms of performance. In particular, it can cost additional learning time and can lower predictive accuracies. 

\paragraph{Contributions.}
We claim that our approach can improve learning performance when learning programs with magic values. 
To support our claim, we make the following contributions:
\begin{enumerate}
\item We introduce a procedure for learning programs in domains with large and potentially infinite numbers of constant symbols.
\item We extend the LFF hypothesis constraints to additionally prune hypotheses with constant symbols.
We prove the optimal soundness of these constraints.
\item We implement our approach in \name, which supports learning recursive programs and predicate invention.
\item We experimentally show on multiple domains (including program synthesis, drug design, and game playing) that our approach can (i) scale to large search spaces with millions of constant symbols, (ii) learn from infinite domains, and (iii) outperform existing systems in terms of predictive accuracies and learning times when learning programs with magic values.
\end{enumerate}

\section{Related Work}

\subsubsection*{Numeric discovery}
Early discovery systems identified relevant numerical values using a fixed set of basic operators, such as linear regression, which combined existing numerical values. 
The search followed a combinatorial design \cite{langley1983,zytkow1987} or was based on beam search guided by heuristics, such as correlation \cite{nordhausen1990} or qualitative proportionality \cite{falkenhainer1986}. 
These systems could rediscover physical laws with magic values. 
However, the class of learnable concepts was limited. 
BACON \cite{langley1983}, for instance, cannot learn disjunctions representing multiple equations. 
Conversely, \name{} can learn recursive programs and perform predicate invention. 
Moreover, \name{} can take as input normal logic program background knowledge and is not restricted to a fixed set of predefined operators. 

\subsubsection*{Symbolic regression}
Symbolic regression searches a space of mathematical expressions, using genetic programming algorithms \cite{augusto2000} or formulating the problem as a mixed integer non-linear program \cite{austel2017}. However, these approaches cannot learn recursive programs nor perform predicate invention and are restricted to learning mathematical expressions.
\subsubsection*{Program synthesis}
Program synthesis \cite{mis} approaches based on the enumeration of the search space \cite{si2019,ellis2021} struggle to learn in domains with a large number of constant symbols. 
For instance, the Apperception engine \cite{evans2021} disallows constant symbols in learned hypotheses, apart from the initial conditions represented as ground facts.  To improve the likelihood of identifying relevant constants, Hemberg et al. \cite{hemberg2019} manually identify a set of constants from the problem description. Compared to generate-and-test approaches, analytical approaches do not enumerate all candidate programs and can be faster \cite{kitzelmann2009inductive}.

Several program synthesis systems consider partial programs in the search. 
Neo \cite{feng2018} constructs partial programs, successively fills their unassigned parts, and prunes partial programs which have no feasible completion.
By contrast, \name{} only fills partial hypotheses with constant symbols. Moreover, \name{} evaluates hypotheses based on logical inference only while Neo also uses statistical inference. Finally, Neo cannot learn recursive programs. Perhaps the most similar work is Sketch \cite{solar2009}, which uses an SAT solver to search for suitable constants given a partial program. This approach expects as input a skeleton of a solution: it is given a partial program and the task is to fill in the magic values with particular constants symbols. Conversely, \name{} learns both the program and the magic values.
\subsubsection*{ILP}


\paragraph{Bottom clauses.}
Early ILP approaches, such as \progol{} \cite{muggleton1995} and \ale{} \cite{srinivasan2001}, use bottom clauses \cite{muggleton1995} to identify magic values.
The bottom clause is the logically most-specific clause that explains an example. 
By constructing the bottom clause, these approaches restrict the search space and, in particular, identify a subset of relevant constant symbols to consider. 
However, this bottom clause approach has multiple limitations.
First, the bottom clause may grow large which inhibits scalability. 
Second, this approach cannot use constants that do not appear in the bottom clause, which is constructed from a single example. 
Third, this approach struggles to learn recursive programs and does not support predicate invention.
Finally, as they rely on mode declarations \cite{muggleton1995} to build the bottom clause, they need to be told which argument of which relations should be bound to a constant.
\paragraph{Lazy evaluation.}
The most related work is an extension of \ale{} that supports \emph{lazy evaluation} \cite{srinivasan1999}. During the construction of the bottom clause, \ale{} replaces constant symbols with existentially quantified output variables.
During the refinement search of the bottom clause, \ale{} finds substitutions for these variables by executing the partial hypothesis on the positive and negative examples. 
In other words, instead of enumerating all constant symbols, lazy evaluation only considers constant symbols computable from the examples. Therefore, lazy evaluation provides better scalability to large domains.
This approach can identify constant symbols not seen in the bottom clause. Moreover, in contrast with \name, it also can identify constant symbols whose value arises from reasoning from multiple examples, such as coefficients in linear regression or numerical inequalities. It also can predict output numerical variables using custom loss functions measuring error \cite{srinivasan2006}.
However, this approach inherits some of the limitations of bottom clause approaches aforementioned including limited learning of recursion and lack of predicate invention.
Moreover, the user needs to provide a definition capable of computing appropriate constant symbols from lists of inputs, such as a definition for computing a threshold or regression coefficients from data. The user must also provide a list of variables that should be lazy evaluated or bound to constant symbols in learned hypotheses. 


\paragraph{Regression.}
First-order regression \cite{karalivc1997} and structural regression tree \cite{kramer1996} predict continuous numerical values from examples and background knowledge. First-order regression builds a logic program that can include literals performing linear regression, whereas \name{} cannot perform linear regression. Structural regression tree builds trees with a numerical value assigned to each leaf. 
In contrast with \name, these two approaches do not learn optimal programs.

\paragraph{Logical decision and clustering trees.}
\til{} \cite{blockeel1998} and TIC \cite{blockeel1998b} are logical extensions of decision tree learners and can learn hypotheses with constant values as part of the nodes that split the examples. 
These nodes are conjunctions built from the mode declarations. \til{} and TIC evaluate each candidate node, and select the one which results in the best split of the examples. \til{} can also use a discretisation procedure to find relevant numerical constants from large, potentially infinite domains, while making the induction process more efficient \cite{blockeel1997}. However, this approach only handles numerical values while \name{} can handle magic values of any type. 
Moreover, \til{} cannot learn recursive programs and struggles to learn from small numbers of examples.

\paragraph{Meta-interpretive learning.}
Meta-interpretive learning (MIL) \cite{muggleton2014} uses meta-rules, which are second-order clauses acting as program templates, to learn programs. 
A MIL learner induces programs by searching for substitutions for the variables in meta-rules. These variables usually denote predicate variables, i.e. variables that can be bound to a predicate symbol. For instance, the MIL learner \metagol{} finds variable substitutions by constructing a proof of the examples. \metagol{} can learn programs with magic values by also allowing some variables in meta-rules to be bound to constant symbols. With this approach, \metagol{}, therefore, never considers constants which do not appear in the proof of at least one positive example and thus does not enumerate all constants in the search space. Our magic value approach is similar in that we construct a hypothesis with variables in it, then find substitutions for these variables by testing the hypothesis on the training examples. However, a key difference is that \metagol{} needs a user-provided set of meta-rules as input to precisely define the structure of a hypothesis, which is often difficult to provide, especially when learning programs with relations of arity greater than two.
Moreover, \metagol{} does not remember failed hypotheses during the search and might consider again hypotheses which have already been proved incomplete or inconsistent. 
Conversely, \name{} can prune the hypothesis space upon failure of completeness or consistency with the examples, which can improve learning performance.

\paragraph{Meta-level ILP.}
To overcome the limitations of older ILP systems, many recent ILP approaches are \emph{meta-level} \cite{cropper2020} approaches, which predominately formulate the ILP problem as a declarative search problem.
A key advantage of these approaches is greater ability to learn recursive and optimal programs.
Many of these recent approaches precompute every possible rule in a hypothesis \cite{aspal,kaminski2018,evans2018,raghothaman2019}. 
For instance, ASPAL \cite{aspal} precomputes every possible rule in a hypothesis space, which means it needs to ground rules with respect to every allowed constant symbol. 
This pure enumeration approach is intractable for domains with large number of constant symbols and impossible for domains with infinite ones. 
Moreover, the variables which should be bound to constants must be provided as part of the mode declarations by the user \cite{aspal}.

Other recent meta-level ILP systems, such as \deltailp{} \cite{evans2018} and \popper{} \cite{cropper2020b}, do not directly allow constant symbols in clauses but instead require that constant symbols are provided as unary predicates. These unary predicates are assumed to be user-provided.
Moreover, since the size of the search space is exponential into the number of predicate symbols, this approach prevents scalability and in particular handling domains with infinite number of constant symbols. 
Conversely, \name{} identifies relevant constant symbols by executing hypotheses over the positive examples, and can scale to infinite domains. 
In addition, it does not express constant symbols with additional predicates and thus can learn shorter hypotheses.

\section{Problem Setting}
\paragraph{Logic preliminaries.}
We assume familiarity with logic programming \cite{lloyd2012} but restate some key terminology. A variable is a string of characters starting with an uppercase letter. A function symbol is a string of characters starting with a lowercase letter. A predicate symbol is a string of characters starting with a lowercase letter. The
arity $n$ of a function or predicate symbol $p$ is the number of arguments it takes. A unary or monadic predicate is a predicate with arity one. A constant symbol is a function symbol with arity zero. A term is a variable or a function symbol of arity $n$ immediately followed by a tuple of $n$ terms. An atom is a tuple $p(t_1, ..., t_n)$, where $p$ is a predicate of arity $n$ and $t_1$, ..., $t_n$ are terms, either variables or constants. An atom is ground if it contains no variables. A literal is an atom or the negation of an atom. A clause is a set of literals. A constraint is a clause without a positive literal. A definite clause is a clause with exactly one positive literal. A program is a set of definite clauses. A substitution $\theta = \{v_1 / t_1, ..., v_n/t_n \}$ is the simultaneous replacement of each variable $v_i$ by its corresponding term $t_i$. A clause $C_1$ subsumes a clause $C_2$ if and only if there exists a substitution $\theta$ such that $C_1 \theta \subseteq C_2$. A program $H_1$ subsumes a program $H_2$, denoted $H_1 \preceq H_2$, if and only if $\forall C_2 \in H_2, \exists C_1 \in H_1$ such that $C_1$ subsumes $C_2$. A program $H_1$ is a specialisation of a program $H_2$ if and only if $H_2 \preceq H_1$. A program $H_1$ is a generalisation of a program $H_2$ if and only if $H_1 \preceq H_2$. 

\subsection{Learning from Failures}
Our problem setting is the learning from failures (LFF) \cite{cropper2020b} setting, which in turn is based upon the learning from entailment setting \cite{muggleton1994}. 
LFF uses hypothesis constraints to restrict the hypothesis space. 
LFF assumes a  meta-language $\cal{L}$, which is a language about hypotheses. 
Hypothesis constraints are expressed in $\cal{L}$. 
A LFF input is defined as:
\begin{definition}
A LFF input is a tuple $(E^{+},E^{-},B,{\cal{H}},C)$ where $E^{+}$ and $E^{-}$ are sets of ground atoms representing positive and negative examples respectively, $B$ is a definite program representing background knowledge, ${\cal{H}}$ is a hypothesis space, and $C$ is a set of hypothesis constraints expressed in the meta-language $\cal{L}$.
\end{definition}

\noindent
Given a set of hypotheses constraints $C$, we say that a hypothesis $H$ is consistent with $C$ if, when written in ${\cal{L}}$, $H$ does not violate any constraint in $C$. We call ${\cal{H}}_{C}$ the subset of $\cal{H}$ consistent with $C$. 
We define a LFF solution:
\begin{definition}

\noindent
Given a LFF input $(E^{+},E^{-},B,{\cal{H}},C)$, a LFF solution is a hypothesis $H \in {\cal{H}}_C$ such that $H$ is complete with respect to $E^+$ ($\forall e \in E^+, B\cup H \models e$) and consistent with respect to $E^-$ ($\forall e \in E^-, B\cup H \not\models e$). \label{solution}
\end{definition}

\noindent
Conversely, given a LFF input, a hypothesis $H$ is \emph{incomplete} when $\exists e \in E^{+}, H \cup B \not\models e$, and is \emph{inconsistent} when $\exists e \in E^{-}, H \cup B \models e$. 

In general, there might be multiple solutions given a LFF input. We associate a cost to each hypothesis and prefer the ones with minimal cost. 
We define an optimal solution:
\begin{definition}
Given a LFF input $(E^{+},E^{-},B,{\cal{H}},C)$ and a cost function \textit{cost} : ${\cal{H}} \rightarrow \mathbb{R}$, a LFF optimal solution $H_1$ is a LFF solution such that, for all LFF solution $H_2$, \emph{$cost(H_1)\leq cost(H_2)$}. 
\end{definition}

\noindent
A common bias is to express the cost as the size of a hypothesis.
In the following, we use this bias, and we measure the size of a hypothesis as the number of literals in it.

\subsubsection*{Constraints}
A hypothesis that is not a solution is called a failure. A LFF learner identifies constraints from failures to restrict the hypothesis space. We distinguish several kinds of failures, among which are the following. 
If a hypothesis is incomplete, a \emph{specialisation} constraint prunes its specialisations, as they are provably also incomplete. 
If a hypothesis is inconsistent, a \emph{generalisation} constraint prunes its generalisations, as they are provably also inconsistent. 
A hypothesis is totally incomplete when $\forall e \in E^{+}, H \cup B \not\models e$. 
If a hypothesis is totally incomplete, a \emph{redundancy constraint} prunes hypotheses that contain one of its specialisations as a subset \cite{cropper2021}.
These constraints are optimally sound: they do not prune optimal solutions from the hypothesis space \cite{cropper2020b}.
\begin{example}[Hypotheses constraints] We call \emph{c2} the unary predicate which holds when its argument is the number 2. Consider the following positive examples $E^+$ and the hypothesis $H_0$: \label{constraint_ex}
\begin{align*}
E^+ = \{f([b,a]), f([c,a,e])\}
\end{align*}
\begin{center}
\begin{tabular}{l}
\emph{$H_0$: f(A) $\leftarrow$ length(A,B), $c_2$(B)}
\end{tabular}
\end{center}
\smallbreak

\noindent The second example is a list of length 3 while the hypothesis $H_0$ only entails lists of length 2. Therefore, the hypothesis $H_0$ does not cover the second positive example and thus is incomplete. We can soundly prune all its specialisations as they also are incomplete. In particular, we can prune the specialisations $H_1$ and $H_2$:\\

\begin{center}
\begin{tabular}{l}
\emph{$H_1$: f(A) $\leftarrow$ length(A,B), $c_2$(B), head(A,B)}\\
\emph{$H_2$: f(A) $\leftarrow$ length(A,B), $c_2$(B), tail(A,C), empty(C)}
\end{tabular}
\end{center}
\end{example}
\section{Magic Evaluation}
\begin{figure}
\begin{center}
\begin{tabular}{l|l}
\emph{$H_{1}$: f(A) $\leftarrow$ length(A,B), $c_1$(B)}& \multirow{4}{*}{\emph{$H$: f(A) $\leftarrow$ length(A,B), @magic(B)}}\\
\emph{$H_{2}$: f(A) $\leftarrow$ length(A,B), $c_2$(B)}&\\
\emph{$H_{3}$: f(A) $\leftarrow$ length(A,B), $c_3$(B)}&\\
\emph{$H_{4}$: f(A) $\leftarrow$ length(A,B), $c_4$(B)}&\\
\emph{$H_{5}$: f(A) $\leftarrow$ length(A,B), $c_5$(B)}&\\
\emph{$H_{6}$: f(A) $\leftarrow$ length(A,B), $c_6$(B)}&\\
... & 
\end{tabular}
\end{center}
\caption{Some hypotheses considered by \popper{} (left) and \name{} (right). $c_1$, $c_2$, $c_3$, $c_4$, $c_5$, and $c_6$ are unary predicates that hold when their argument is the number 1, 2, 3, 4, 5, or 6 respectively. These unary predicates are assumed to be user-provided.}
\label{hypotheses}
\end{figure}
The constraints described in the previous section prune hypotheses. 
In particular, they can prune hypotheses with constant symbols as shown in Example \ref{constraint_ex}. However, hypotheses identical but with different constant symbols are treated independently despite their similarities. 

For instance, \popper{} could consider all of the hypotheses represented on the left of Figure \ref{hypotheses}. 
Each of these hypotheses would be considered independently. For each of them, \popper{} learns constraints which prune specialisations, generalisations, or redundancy of this single hypothesis but do not apply to other hypotheses.
By contrast, as shown on the right of Figure \ref{hypotheses}, \name{} represents all these hypotheses jointly as a single one by using variables in place of constant symbols. 
Thus, \name{} reasons simultaneously about hypotheses with similar program structure but different constant symbols.

\name{} extends specialisation, generalisation, and redundancy constraints to apply to such partial hypotheses. 

Moreover, the unary predicate symbols used by \popper{} must be provided as bias: it is assumed the user can provide a finite and tractable number of them. Conversely, \name{} represents the set of hypotheses with similar structure but with different constant symbols as a single one, and therefore can handle infinite constant domains.

In this section, we introduce \name{}'s representation, present these extended constraints, and prove they are optimally sound.
\subsection{Magic Variables}
A LFF learner uses a meta-language $\cal{L}$ to reason about hypotheses. We extend this meta-language $\cal{L}$ to represent partial hypotheses with unbound constant symbols. We define a magic variable:
\begin{definition}
A magic variable is an existentially quantified first-order variable.
\end{definition}
\noindent
A magic variable is a placeholder for a constant symbol. It marks a variable as a constant but does not require the particular constant symbol to be identified. 
Particular constant symbols can be identified in a latter stage.
We represent magic variables with the unary predicate symbol \emph{@magic}. For example, in the following program $H$, the variable $B$ marked with the syntax \emph{@magic} is a magic variable:

\begin{center}
\begin{tabular}{l}
$H$: \emph{f(A) $\leftarrow$ length(A,B), @magic(B)}\\
\end{tabular}
\end{center}

This magic variable is not yet bound to any particular value. The use of the predicate symbol \emph{@magic} allows us to concisely represent the set of all possible substitutions of a variable.

 The predicate symbol \emph{@magic} is an internal predicate. 
For this reason, literals with this predicate symbol are not taken into account in the rule size. For instance, the hypothesis $H$ above has size 2. 
Therefore, compared to approaches that use additional unary body literals to identify constant symbols, our representation represents hypotheses with constant symbols with fewer literals.
\subsection{Magic Hypotheses}
A \emph{magic hypothesis} is a hypothesis with at least one magic variable. An \emph{instantiated hypothesis}, or \emph{instantiation}, is the result of substituting magic variables with constant symbols in a magic hypothesis. 
\emph{Magic evaluation} is the process of identifying a relevant subset of substitutions for magic variables in a magic hypothesis to form instantiations.


\begin{example}[Magic hypothesis]
The magic hypothesis $H$ above may have the following corresponding instantiated hypotheses, or instantiations, $I_1$ and $I_2$:
\begin{center}
\begin{tabular}{l}
$I_1$ : \emph{f(A) $\leftarrow$ length(A,2)}\\
$I_2$ : \emph{f(A) $\leftarrow$ length(A,0)}
\end{tabular}
\end{center}
\end{example}

\noindent
Magic hypotheses allow us to represent the hypothesis space more compactly and to reason about the set of all instantiations of a magic hypothesis simultaneously. 
For instance, the magic hypothesis $H$ above represents concisely all its instantiations, including $I_1$ and $I_2$, amongst many other ones. The only instantiation of a non-magic hypothesis is itself.

In practice, we are not interested in all instantiations of a magic hypothesis, but only in a subset of relevant instantiations. 
In the following, we consider a magic evaluation procedure which only considers instantiations that, together with the background knowledge, entail at least one positive example. We show we can ignore other instantiations. 
\subsection{Constraints}\label{constraints}

To improve learning performance, we prune the hypothesis space with constraints \cite{cropper2020b}.
Given our hypothesis representation, each constraint prunes a set of magic hypotheses, each of which represents the set of its instantiations. 
In other words, for each magic hypothesis pruned, we eliminate all its instantiations.
We identify constraints that are optimally sound in that they do not eliminate optimal solutions from the hypothesis space. 
Specifically, we consider extensions of specialisation, generalisation, and redundancy constraints for magic hypotheses. 
We describe them in turn. 
The proofs are in the appendix.

\subsubsection{Extended Specialisation Constraint}
We first extend specialisation constraints. 
If all the instantiations of a magic hypothesis, together with the background knowledge, entail at least one positive example and are incomplete, then all specialisations of this hypothesis are incomplete:

\begin{proposition}[Extended specialisation constraint]
Let $(E^{+},E^{-},B,{\cal{H}},C)$ be a LFF input, $H_1 \in {\cal{H}}_C$, and $H_2 \in {\cal{H}}_C$ be two magic hypotheses such that $H_1 \preceq H_2$. 
If all instantiation $I_1$ of $H_1$ such that $\exists e \in E^{+}, B\cup I_1 \models e$ are incomplete, then all instantiation of $H_2$ also are incomplete. \label{prop2}
\end{proposition}

\noindent
We provide an example to illustrate this proposition.
\begin{example}[Extended specialisation constraint] \label{example_spec} Consider the dyadic predicate \emph{head} which takes as input a list and returns its first element. Consider the following positive examples $E^+$ and the magic hypothesis $H_0$:
\begin{align*}
E^+ = \{f([b,a]), f([c,a,e])\}
\end{align*}
\begin{center}
\begin{tabular}{l}
\emph{$H_0$: f(A) $\leftarrow$ head(A,B), @magic(B)}
\end{tabular}
\end{center}

This hypothesis holds for lists whose first element is a particular constant symbol to be determined. This hypothesis $H_0$ has the following two instantiations $I_{0,1}$ and $I_{0,2}$ covering at least one positive example:
\begin{center}
\begin{tabular}{l}
\emph{$I_{0,1}$: f(A) $\leftarrow$ head(A,b)}\\
\emph{$I_{0,2}$: f(A) $\leftarrow$ head(A,c)}
\end{tabular}
\end{center}

The first instantiation $I_{0,1}$ holds for lists whose head is the element \emph{b}. This instantiation covers the first positive example. The second instantiation $I_{0,2}$ holds for lists whose head is the element \emph{c}. It covers the second positive example.
However, each of these instantiations is incomplete and too specific. Therefore, no instantiation of $H_0$ can entail all the positive examples. As such, all specialisations of $H_0$ can be pruned, including magic hypotheses such as $H_1$ and $H_2$:
\begin{center}
\begin{tabular}{l}
\emph{$H_1$: f(A) $\leftarrow$ head(A,B), @magic(B), odd(B)}\\
\emph{$H_2$: f(A) $\leftarrow$ head(A,B), @magic(B), tail(A,C), head(C,D), @magic(D)}
\end{tabular}
\end{center}
\end{example}

\subsubsection{Extended Generalisation Constraint}
We now extend generalisation constraints. 
If all the instantiations of a magic hypothesis together with the background knowledge entail at least one positive example are inconsistent, then we can prune non-recursive generalisations of this hypothesis and they are either inconsistent or non-optimal:
\begin{proposition}[Extended generalisation constraint]
Let $(E^{+},E^{-},B,{\cal{H}},C)$ be a LFF input, $H_1 \in {\cal{H}}_C$ and $H_2 \in {\cal{H}}_C$ be two magic hypotheses such that $H_2$ is non-recursive and $H_2 \preceq H_1$. If all instantiation $I_1$ of $H_1$ such that $\exists e \in E^{+}, B\cup I_1 \models e$ are inconsistent, then all instantiations of $H_2$ are inconsistent or non-optimal. \label{prop3}
\end{proposition}
\noindent
We illustrate generalisation constraints with the following example and give a counter-example to explain why non-recursive hypotheses cannot be pruned.
\begin{example}[Extended generalisation constraint] Consider the following positive examples $E^+$, the negative examples $E^-$ and the magic hypothesis $H_0$:
\begin{align*}
&E^+ = \{f([b,a]), f([c,a,e])\}\\
&E^- = \{f([b]), f([c])\}
\end{align*}
\begin{center}
\begin{tabular}{l}
\emph{$H_0$: f(A) $\leftarrow$ head(A,B), @magic(B)}
\end{tabular}
\end{center}
This hypothesis $H_0$ has the following two instantiations $I_{0,1}$ and $I_{0,2}$ covering at least one of these positive examples:
\begin{center}
\begin{tabular}{l}
\emph{$I_{0,1}$: f(A) $\leftarrow$ head(A,b)}\\
\emph{$I_{0,2}$: f(A) $\leftarrow$ head(A,c)}
\end{tabular}
\end{center}
The first instantiation $I_{0,1}$ holds for lists whose head is the element \emph{b}. This instantiation covers the first positive example and the first negative example. The second instantiation $I_{0,2}$ holds for lists whose head is the element \emph{c}. It covers the second positive example and the second negative example. Each of these instantiations is inconsistent and thus is too general. As such, all non-recursive generalisations of $H_0$ can be pruned. In particular, the magic hypotheses $H_1$ and $H_2$ below are non-recursive generalisations of $H_0$ and can be pruned:
\begin{align*}
&H_1: 
\left\{
    \begin{array}{l}
    \emph{f(A) $\leftarrow$ head(A,B), @magic(B)}\\
    \emph{f(A) $\leftarrow$ length(A,B), @magic(B)}
    \end{array}
\right.\\
&H_2: \left\{
    \begin{array}{l}
    \emph{f(A) $\leftarrow$ head(A,B), @magic(B)}\\
    \emph{f(A) $\leftarrow$ head(A,B), negative(B)}
    \end{array}
\right.
\end{align*}
However, there might exist other instantiations of $H_0$ which do not cover any positive examples but are not inconsistent, such as $I_{0,3}$:
\begin{center}
\begin{tabular}{l}
\emph{$I_{0,3}$: f(A) $\leftarrow$ head(A,a)}
\end{tabular}
\end{center}
This instantiation could be used to construct a recursive solution, such as $I$:
\begin{center}
$$
I: \left\{
    \begin{array}{l}
    \emph{f(A) $\leftarrow$ head(A,a)}\\
    \emph{f(A) $\leftarrow$ tail(A,B), f(B)}
    \end{array}
\right.
$$
\end{center}
The instantiation $I$ holds for list which contain the element \emph{a} at any position.
\end{example}

\subsubsection{Extended Redundancy Constraint}
We extend redundancy constraints for magic hypotheses. If a magic hypothesis has no instantiations which, together with the background knowledge, entail at least one positive example, we show that it is redundant when included in any non-recursive hypothesis.
\begin{proposition}[Extended redundancy constraint]
Let $(E^{+},E^{-},B,{\cal{H}},C)$ be a LFF input, $H_1 \in {\cal{H}}_C$ be a magic hypothesis. If $H_1$ has no instantiation $I_1$ such that $\exists e \in E^{+}, B\cup I_1 \models e$, then all non-recursive magic hypotheses $H_2$ which contain a specialisation of $H_1$ as a subset are non-optimal. \label{prop1}
\end{proposition}


\noindent
We illustrate this proposition with the following example and provide a counter-example to explain why non-recursive hypotheses cannot be pruned.
\begin{example}[Extended redundancy constraint] Consider the following positive examples $E^+$ and the magic hypothesis $H_0$:
\begin{align*}
E^+ = \{f([b,c]), f([f,g,c])\}
\end{align*}
\begin{center}
\begin{tabular}{l}
\emph{H$_0$: f(A) $\leftarrow$ head(A,B), @magic(B), tail(A,C),empty(C)}
\end{tabular}
\end{center}
This hypothesis $H_0$ holds for lists which contain a single element which is a particular constant to be determined. However, both examples have length 2. Therefore, among the possible instantiations of the hypothesis $H_0$, there are no instantiations which, together with the background knowledge, cover at least one positive example. $H_0$ cannot entail any of the positive examples and is redundant when included in a non-recursive hypothesis. As such, all hypotheses which contain a specialisation of $H_0$ as a subset are non-optimal. 
In particular, the magic hypotheses $H_1$ and $H_2$ below contain a specialisation of $H_0$ as a subset. They are non-optimal and can be pruned:
\begin{align*}
&H_1: 
\left\{
    \begin{array}{l}
    \emph{f(A) $\leftarrow$ head(A,B), @magic(B), odd(B), tail(A,C), empty(C)}\\
    \emph{f(A) $\leftarrow$ length(A,B), @magic(B)}
    \end{array}
\right. \\
&H_2: \left\{
    \begin{array}{l}
    \emph{f(A) $\leftarrow$ head(A,B), @magic(B), tail(A,C),empty(C), length(A,D), odd(D)}\\
    \emph{f(A) $\leftarrow$ head(A,B), negative(B)}
    \end{array}
\right.
\end{align*}
However, there might exist other instantiations of $H_0$ which are not redundant in recursive hypotheses. For instance, the following recursive instantiated hypothesis $I$ includes an instantiation of $H_0$ as a subset but may not be non-optimal:
\begin{align*}
I: \left\{
    \begin{array}{l}
    \emph{f(A) $\leftarrow$ head(A,c), tail(A,C),empty(C)}\\
    \emph{f(A) $\leftarrow$ tail(A,B), f(B)}
    \end{array}
\right.
\end{align*}
This instantiation holds for lists whose last element is the constant \emph{c}. It covers both positive examples but none of the negative examples.
\end{example}
\noindent
While extended specialisation constraints are sound, extended redundancy and generalisation constraints are only optimally sound. They might prune solutions from the hypothesis space but do not prune optimal solutions.
\subsubsection{Constraint summary}
We summarise our constraint framework as follows. Given a magic hypothesis $H$, the learner can infer the following extended constraints under the following conditions: \label{constraint_framework}
\begin{enumerate}
    \item If all instantiations of $H$ which, together with the background knowledge, entail at least one positive example are incomplete, according to Proposition \ref{prop2}, we can prune all its specialisations.
    \item If all instantiations of $H$ which, together with the background knowledge, entail at least one positive example are inconsistent, according to Proposition \ref{prop3}, we can prune all its non-recursive generalisations.
    \item If the magic hypothesis $H$ has no instantiation which, together with the background knowledge, entail at least one positive example, according to Proposition \ref{prop1}, we can prune all non-recursive hypotheses which contain one of its specialisations as a subset.
\end{enumerate}
While Proposition \ref{prop2} can prune recursive hypotheses, Proposition \ref{prop3} and Proposition \ref{prop1} do not prune recursive hypotheses. Therefore, pruning is stronger when recursion is disabled.

\noindent
We have described our representation of the hypothesis space with magic hypotheses. We have extended specialisation, generalisation, and redundancy constraints to prune magic hypotheses and we have demonstrated these extended constraints are optimally sound. The next section theoretically evaluates the gain over the size of the search space of using magic hypotheses compared to identifying constant symbols with unary predicates.

\subsection{Theoretical Analysis\label{theoretical}}
Our representation includes magic hypotheses which contain magic variables. Each magic variable stands for the set of its substitutions. Therefore, we do not enumerate constant symbols in the hypothesis space by opposition with existing approach. Our experiments focus on comparing \name{} with approaches which enumerate possible constant symbols with unary body predicates. We focus in this section on theoretically evaluating the reduction over the hypothesis space size of not enumerating all candidate constant symbols as unary predicates, and instead using magic variables. 
\begin{proposition}
Let $D_b$ be the number of body predicates available in the search space, $m$ be the maximum number of body literals allowed in a clause, $c$ the number of constant symbols available, and $n$ the maximum number of clauses allowed in a hypothesis. Then the maximum number of hypotheses in the hypothesis space can be multiplied by a factor of $(\frac{D_b+c}{D_b})^{mn}$ if representing constants with unary predicate symbols, one per allowed constant symbol, compared to using magic variables. \label{prop:gain}
\end{proposition}
\noindent
 A proof of Proposition \ref{prop:gain} is in the appendix. Proposition \ref{prop:gain} shows that allowing magic variables can reduce the size of the hypothesis space compared to enumerating constant symbols through unary predicate symbols. The ratio is a increasing function of the number of constant symbols available and the complexity of hypotheses, measured as the number of clauses allowed in hypotheses $n$ and the number of body literals allowed in clauses $m$. Similar analysis can be conducted for approaches which enumerate constant symbols in the arguments of clauses. More generally, Proposition \ref{prop:gain} suggests that enumerating constant symbols can increase the size of the hypothesis space compared to using magic variables.

\section{Implementation}
We now describe \name{}, which implements our magic evaluation idea.
We first describe \popper{}, on which \name{} is based. 
\subsection{\popper}
\popper{} \cite{cropper2020b} is a LFF learner. It takes as input a LFF input, which contains a set of positive ($E^+$) and negative ($E^-$) examples, background knowledge ($B$), a bound over the size of hypotheses allowed in ${\cal{H}}$, and a set of hypotheses constraints ($C$). 
\popper{} represents hypotheses in a meta-language $\cal{L}$. 
This meta-language $\cal{L}$ contains literals  \emph{head$\_$literal/4} and \emph{body$\_$literal/4} representing head and body literals respectively. These literals have arguments \emph{(Clause,Pred,Arity,Vars)} and denote that there is a head or body literal in the clause \emph{Clause}, with the predicate symbol \emph{Pred}, arity \emph{Arity}, and variables \emph{Vars}.
For instance, the following set of literals:
\begin{center}
\emph{\{head$\_$literal(0,empty,1,(0)), body$\_$literal(0,length,2,(0,1)), body$\_$literal(0,zero,1,(1))\}}
\end{center}
represents the following clause with id 0:
\begin{center}
\begin{tabular}{l}
\emph{empty(A) $\leftarrow$ length(A,B), zero(B)}
\end{tabular}
\end{center}

\noindent
To generate hypotheses, \popper{} uses an ASP program $P$ whose models are hypothesis solutions represented in the meta-language $\cal{L}$.
In other words, each model (answer set) of $P$ represents a hypothesis. A simplified version of the base ASP program (without the predicate declarations which are problem specific) is represented in Figure \ref{fig:asp_encoding}.
\popper{} uses a \emph{generate}, \emph{test}, and \emph{constrain} loop to find a solution. 
First, it generates a hypothesis as a solution to the ASP program $P$ with the ASP system Clingo \cite{gebser2014}. 
\popper{} searches for hypotheses by increasing size, the size being evaluated as the number of literals in a hypothesis. 
\popper{} tests this hypothesis against the examples, typically using Prolog.
If the hypothesis is a solution, it is returned. Otherwise, the hypothesis is a failure: \popper{} identifies the kind of failure and builds constraints accordingly. For instance, if the hypothesis is inconsistent (entails a negative example) \popper{} builds a generalisation constraint. 
\popper{} adds these constraints to the ASP program $P$ to constrain the subsequent \textit{generate} steps. 
This loop repeats until a hypothesis solution is found or until there are no more models to the ASP program $P$.

\begin{figure}
\begin{tabular}{l}
\texttt{head$\_$literal(C,P,A,Vars):-}\\ \texttt{    clause(C),}\\
\texttt{    head$\_$pred(P,A),}\\
\texttt{    vars(A,Vars).}\\
\\
\texttt{0\{body$\_$literal(C,P,A,Vars): clause(C),body$\_$pred(P,A),vars(A,Vars)\}N :-}\\
\texttt{    max$\_$body(N).}\\
\\
\texttt{size(N):-}\\
\texttt{    max$\_$size(MaxSize),}\\
\texttt{    N = 1..MaxSize,}\\
\texttt{    \#sum\{K+1,C : body$\_$size(C,K)\} == N.}\\
\\
\texttt{var(0..N-1):- max$\_$vars(N).}
\\
\texttt{vars(1,(Var1,)):- var(Var1).}\\
\texttt{vars(2,(Var1,Var2)):- var(Var1),var(Var2).}\\
\texttt{vars(3,(Var1,Var2,Var3)):- var(Var1),var(Var2),var(Var3).}\\
\\
\textbf{\texttt{0 \{magic$\_$literal(C,Vars): clause(C),vars(1,Vars)\} M:-
    max$\_$magic(M).}}\\
    \\
\end{tabular}
    \caption{Simplified \popper{} ASP base program for single clause programs. There is exactly one head literal per clause. There are at most N body literals per clause, where N is a user-provided parameter describing the maximum number of body literals allowed in a clause. Our modification is highlighted in bold: we allow at most N variables to be magic variables, where N is a user-provided parameter.}
    
    \label{fig:asp_encoding}
\end{figure}

\subsection{\name}
\name{} builds on \popper{} to support magic evaluation.
\name{} likewise follows a \emph{generate}, \emph{test}, and \emph{constrain} loop to find a solution. 
We describe in turn how each of these steps works.

\paragraph{Generate}\label{generate}
Figure \ref{fig:asp_encoding} shows our modification to \popper's base ASP encoding in bold. In addition to \emph{head$\_$literal/4} and \emph{body$\_$literal/4}, \name{} can express  \emph{magic$\_$literal/2}. Magic literals have arguments \emph{(Clause,Var)} and denote that the variable Var in the clause \emph{Clause} is a magic variable. There can be at most $M$ magic literal in a clause, where $M$ is a user defined parameter with default value 4. This setting expresses the trade-off between search complexity and expressivity.

In addition to the standard \popper{} input, and a maximum number of magic values per clause, \name{} can receive information about which variables can be magic variables. This information can be provided with three different settings: \emph{Arguments}, \emph{Types}, and \emph{All}. For instance, given the predicate declarations represented in Figure \ref{predicate_declarations}, Figure \ref{settings} illustrates how the user can provide additional bias with each of these settings. A user can specify individually a list of some arguments of some predicates (\emph{Arguments}) or a list of variable types (\emph{Types}). Otherwise, if no information is given, \name{} treats any variable as a potential magic variable (\emph{All}). For any of these settings, \name{} searches for a subset of the variables specified by the user for the magic variables. Therefore, \emph{All} always considers a larger hypothesis space than \emph{Arguments} and \emph{Types}. \emph{Arguments} is the setting closest to mode declarations \cite{muggleton1995,blockeel1998,srinivasan2001,aspal}. 
Mode declarations however impose a stricter bias: while \emph{Arguments} treats the flagged arguments as potential magic values, mode declarations specify an exact list of arguments which must be constant symbols\footnote{Forcing variables to be magic variables is not a setting currently available in \name.}. 
With the \emph{All} setting, \name{} can automatically identify which variable to treat as magic variables at the expense of more search. In Section \ref{bias}, we experimentally evaluate the impact on learning performance of these different settings. In Section \ref{q5}, we evaluate the impact on learning performance of allowing magic values (\emph{All} setting) when it is unnecessary.

\begin{figure}
\small
\begin{center}
\begin{tabular}{l|l|l}
\textbf{Predicate} & \textbf{Type} & \textbf{Directions}\\ \hline
\texttt{head$\_$pred(f,1).} & \texttt{(state)} & \texttt{(in)}\\
\texttt{body$\_$pred(cell,4).} & \texttt{(state,pos,color,type)} & \texttt{(in,out,out,out)}\\
\texttt{body$\_$pred(distance,3).} & \texttt{(pos,pos,int)} & \texttt{(in,in,out)}\\
\end{tabular}
\caption{Predicate declarations for the \emph{krk} task. The task is to learn a hypothesis to describe that the white king protects the white rook in the chess endgame king-rook-king.} \label{predicate_declarations}
\end{center}
\end{figure}
\begin{figure}
\small
\begin{center}
\begin{tabular}{l|l|l}
\textbf{Setting} & \textbf{Bias} & \textbf{Example hypothesis}\\ \hline
\multirow{2}{*}{\emph{Arguments}}& cell, 3 & \multirow{2}{*}{f(A)$\leftarrow$ cell(A,B,\textbf{C},D),cell(A,E,\textbf{F},D),distance(B,E,\textbf{H})}\\
& distance, 3&\\ \hline

\multirow{2}{*}{\emph{Types}}& integer & \multirow{2}{*}{f(A)$\leftarrow$ cell(A,B,C,\textbf{D}),cell(A,E,C,\textbf{G}),distance(B,E,\textbf{H})}\\
& type &\\\hline

\emph{All}&  & f(\textbf{A})$\leftarrow$ cell(\textbf{A},\textbf{B},\textbf{C},\textbf{D}),cell(\textbf{A},\textbf{E},\textbf{F},\textbf{G}),distance(\textbf{B},\textbf{E},\textbf{H})\\
\end{tabular}
\end{center}
\caption{Example of the different bias settings for \name{}. Variables that can be magic variables are represented in bold. \emph{Arguments} can treat as magic variables some specified arguments of specified predicates. \emph{Type} can treat as a magic variable any variable of the specified types. \emph{All} expects no additional information and may treat any variable as a magic variable.}
\label{settings}
\end{figure}

The output of the \textit{generate} step is a hypothesis which may contain magic variables, such as the one shown on the right of Figure \ref{hypotheses}. 
By contrast, most ILP approaches \cite{aspal,evans2018,cropper2020b} cannot generate hypotheses with magic variables but instead require enumerating constant symbols. \popper{} and \deltailp{} use unary predicates to represent constant symbols, as shown on the left of Figure \ref{hypotheses}. \aspal{} precomputes all possible rules with some arguments grounded to constant symbols. Conversely, owing to the use of magic variables, \name{} benefits from a more compact representation of the hypothesis space.



\paragraph{Test} 
Magic evaluation is executed during the \textit{test} step. 
To identify substitutions for magic variables, we add magic variables as new head arguments.
We execute the resulting program on the positive examples. We save the substitutions for the new head variables. We then bound these substitutions to their corresponding magic variables and remove the additional head arguments.
\begin{example}[Magic evaluation]
Consider the magic hypothesis $H_1$ below:
\begin{center}
\begin{tabular}{l}
$H_1$: \emph{f(A) $\leftarrow$ length(A,B), @magic(B)}
\end{tabular}
\end{center}
We add magic variables as new head variables. $H$ thus becomes $H_1^\prime$.
\begin{center}
\begin{tabular}{l}
$H_1^\prime$: \emph{f(A,B) $\leftarrow$ length(A,B), @magic(B)}\\
\end{tabular}
\end{center}
 We execute $H_1^\prime$ on the positive examples to find substitutions for the magic variable $B$. Assume the single positive example $f([a,b,c])$. We transform it into $f([a,b,c],B)$ and we find the substitution $3$ for the variable $B$. We bind this value to the magic variable in the hypothesis, which results in the following instantiation:
 \begin{center}
\begin{tabular}{l}
$H_1^\prime$: \emph{f(A,B) $\leftarrow$ length(A,3)}\\
\end{tabular}
\end{center}
\end{example}
\begin{example}[Magic evaluation of recursive hypothesis]
Similarly, the recursive hypothesis $H_2$ below becomes $H_2^{\prime}$.
\begin{align*}
&H_2: \left\{
    \begin{array}{l}
    \emph{f(A) $\leftarrow$ length(A,B), @magic(B)}\\
    \emph{f(A) $\leftarrow$ head(A,B), @magic(B), tail(A,C), f(C)}
    \end{array}
\right.\\
&H_2^{\prime}: \left\{
    \begin{array}{l}
    \emph{f(A,B,D) $\leftarrow$ length(A,B), @magic(B)}\\
    \emph{f(A,B,D) $\leftarrow$ head(A,D), @magic(D), tail(A,C), f(C,B,D)}
    \end{array}
\right.
\end{align*}
We execute $H_2^\prime$ on the positive examples to find substitutions for the magic variables $B$ and $D$.
\end{example}
\noindent
With this procedure, \name{} only identifies constants which can be obtained from the positive examples. In this sense, \name{} does not consider irrelevant constant symbols. 
\begin{example}[Relevant instantiations]
Given the positive examples $E^+~=~\{f([a,e]),f([])\}$, we consider only the two instantiations $I_{1,1}$ and $I_{1,2}$ for the magic hypothesis $H_1$:
\begin{center}
\begin{tabular}{l}
$H_1$: \emph{f(A) $\leftarrow$ length(A,B), @magic(B)}\\
\end{tabular}
\begin{tabular}{l}
\emph{$I_{1,1}$: f(A) $\leftarrow$ length(A,2)}\\
\emph{$I_{1,2}$: f(A) $\leftarrow$ length(A,0)}
\end{tabular}
\end{center}
\end{example}
\noindent
We use Prolog to execute programs because of its ability to use lists and handle large, potentially infinite, domains. As a consequence of using Prolog, our reasoning to deduce candidate magic values is based on backward chaining, in contrast to systems that rely on forward chaining \cite{aspal,evans2018,kaminski2018,evans2021}.

\label{section:separable}
A limitation of the aforementioned approach is the execution time of learned programs to identify all possible bindings. This approach is especially expensive when a hypothesis contains multiple magic variables, in which case one must consider the combinations of their possible bindings. 

\begin{example}[Execution time complexity] Consider the hypothesis $H$: \label{separable}
\begin{align*}
H: \left\{
    \begin{array}{l}
    \emph{f(A) $\leftarrow$ member(A,B1),@magic(B1)}\\
    \emph{f(A) $\leftarrow$ member(A,B2),@magic(B2)}\\
    \end{array}
\right.
\end{align*}
The hypothesis $H$ is the disjunction of two clauses, each of which contains one magic value, respectively $B1$ and $B2$. Since $B1$ and $B2$ can be bound to different constant symbols, this hypothesis is allowed in the search space despite having two clauses with the exact same literals. More generally, we allow identical clauses with magic variables.

This hypothesis means that any of two particular elements appears in a list. We search for substitutions for the magic variables $B1$ and $B2$. We call $n$ the size of input lists. The number of substitutions for the magic variable $B1$ in the first clause is $O(n)$. Similarly, the number of substitutions for the magic variable $B2$ in the second clause is $O(n)$. Therefore, the number of instantiations for $H$ is $O(n^2)$.
\end{example}



\paragraph{Constrain} If \name{} identifies that a hypothesis has no instantiation, no complete instantiation, or no consistent instantiation, it generates constraints as explained in Section \ref{constraint_framework}. 
Additionally, \name{} generates a banish constraint if no other constraints can be inferred. The banish constraint prunes this single hypothesis from the hypothesis space.
In other words, it ensures that the same hypothesis will not be generated again in subsequent \emph{generate} steps \cite{cropper2020b}. 
These constraints prune the hypothesis space and constrain the following iterations.


\section{Experiments}

We now evaluate our approach.

\subsection{Experimental design}

Our main claim is that \name{} can improve learning performance compared to current ILP systems when learning programs with magic values. 
Our experiments, therefore, aim to answer the question:


\begin{enumerate}
\item[\textbf{Q1}] How well does \name{} perform compared to other approaches?
\end{enumerate}

\noindent
To answer \textbf{Q1}, we compare \name{} against \metagol, \ale{}, and \popper{}\footnote{We also considered \textsc{ASPAL} \cite{aspal}. However, as it precomputes every possible rule in a hypothesis, it does not scale to our experimental domains.}. 
\name{} uses different biases than \metagol{} and \ale{}. Therefore a direct comparison is difficult and our results should be interpreted as indicative only.
By contrast, as \name{} is based on \popper{}, the comparison against \popper{} is more controlled.
The experimental difference between the two is the addition of our magic evaluation procedure and the use of extended constraints. 

A key limitation of approaches that enumerate all possible constants allowed in a rule \cite{aspal,evans2018,cropper2020b} is difficulty learning programs from infinite domains.
By contrast, we claim that \name{} can learn in infinite domains.
Therefore, our experiments aim to answer the question:
\begin{enumerate}
\item[\textbf{Q2}] Can \name{} learn in infinite domains?
\end{enumerate}

\noindent
To answer \textbf{Q2}, we consider several tasks in infinite and continuous domains that require magic values as real numbers or integers. 

Proposition \ref{prop:gain} shows that our magic evaluation procedure can reduce the search space and thus improve learning performance compared to using unary body predicates. We thus claim that \name{} can improve scalability compared to \popper. 
To explore this claim, our experiments aim to answer the question:
\begin{enumerate}
\item[\textbf{Q3}] How well does \name{} scale?
\end{enumerate}

\noindent
To answer \textbf{Q3}, we vary the number of (i) constant symbols in the background knowledge, (ii) magic values in the target hypotheses, and (iii) training examples. We use as baseline \popper{}. We compare our experimental results with our theoretical analysis from Section \ref{theoretical}. 


Unlike existing approaches, \name{} does not need to be told which variables may be magic variables but can automatically identify this information. However, it can use this information if provided by a user.
To evaluate the importance of this additional information, our experiments aim to answer the question:
\begin{enumerate}
\item[\textbf{Q4}] What effect does additional bias about magic variables have on the learning performance of \name{}? 
\end{enumerate}

\noindent
To investigate \textbf{Q4}, we compare different settings for \name{}, each of which assumes different information regarding which variables may be magic variables. We use as baseline \popper{}.

Our approach should improve learning performance when learning programs with magic values. However, in practical applications, it is unknown whether magic values are necessary. To evaluate the cost in performance when magic values are unnecessary, our experiments aim to answer the question:
\begin{enumerate}
\item[\textbf{Q5}] What effect does allowing magic values have on the learning performance when magic values are unnecessary?
\end{enumerate}
To answer \textbf{Q5}, we compare the learning performance of \name{} and \popper{} on problems that should not require magic values. We set \name{} to allow any variable to potentially be a magic value.
\subsubsection{Experimental settings}
Given $p$ positive and $n$ negative examples, $tp$ true positives and $tn$ true negatives, we define the predictive accuracy as $\frac{(\frac{tp}{p}+\frac{tn}{n})}{2}$. We measure mean predictive accuracies, mean learning times, and standard errors of the mean over 10 repetitions. We use an 8-Core 3.2 GHz Apple M1 and a single CPU\footnote{The experimental data and code for reproducing the experiments are available at https://github.com/celinehocquette/magicpopper.git}.

\subsubsection{Systems settings}
\paragraph{\ale{}.}
\ale{} is allowed constant symbols through the mode declarations or lazy evaluation.

\paragraph{\metagol{}.}
\metagol{} needs as input second-order clauses called metarules. We provide \metagol{} with a set of almost universal metarules for a singleton-free fragment of monadic and dyadic Datalog \cite{cropper2020e} and additional \textit{curry} metarules to identify constant symbols as existentially quantified first-order variables. 

\paragraph{\popper{} and \name{}.}
Both systems use \popper{} 2.0.0 (also known as \popperplus{}) \cite{popper+}.
We provide \popper{} with one unary predicate symbol for each constant symbol available in the background knowledge. We set for both systems the same parameters bounding the search space (maximum number of variables and maximum number of literals in the body of clauses). Therefore, since \name{} does not count magic literals in program sizes, it considers a larger search space than \popper. We provide both systems with types for predicate symbols. In particular, unary predicates provided to \popper{} are typed. Therefore, to ensure a fair comparison, we provide \name{} with a list of types to describe the set of variables which may be magic variables. As explained in Section \ref{generate}, we could have instead provided \name{} with a list of arguments of particular predicate symbols to describe the set of variables which may be magic variables. Providing a list of predicate arguments would have been a setting closer to mode declarations, which \ale{} uses. However, when specifying types for magic variables, the search space is larger than when specifying particular arguments of some predicates symbols. Moreover, our setting specifies which variables can be magic variables, and \name{} searches for a subset of these variables. Conversely, modes specify which variables must be constant symbols. In this sense, this setting for \name{} considers a larger hypothesis space than \ale. In Section \ref{bias}, we evaluate and compare the effect on learning performance of these different settings for specifying magic variables.


  

\subsection{\textbf{Q1}: comparison with other systems}
\subsubsection{Experimental domains} \label{domains}
We compare \name{} against state-of-the-art ILP systems. This experiment aims to answer \textbf{Q1}. We consider several domains. Full descriptions of these domains are in the appendix. 
We use a timeout of 600s for each task.
\paragraph{IGGP.} In inductive general game playing (IGGP) \cite{cropper2020a}, agents are given game traces from the general game playing competition \cite{genesereth2013}. The task is to induce a set of game rules that could have produced these traces. We use four IGGP games which contain constant symbols: \textit{md} (minimal decay), \textit{buttons},  \textit{coins}, and \textit{gt-centipede}. We learn the \textit{next} relation in each game, the \textit{goal} relation for \textit{buttons}, \textit{coins}, \textit{gt-centipede} and the \textit{legal} relation for \textit{gt-centipede}. These tasks involve the identification of respectively 5, 31, 3, 14, 6, 4, 29 and 8 magic values. Figures \ref{fig:md} and \ref{fig:buttons} represent examples of some target hypotheses. We measure balanced accuracies and learning times.

\begin{figure}[t]
\begin{tabular}{l}
\emph{next\_val(A,\textbf{5}) $\leftarrow$ does(A,\textbf{player},\textbf{press\_button)}}\\
\emph{next\_val(A,B) $\leftarrow$ does(A,\textbf{player},\textbf{noop}), true\_val(A,C), succ(B,C)}
\end{tabular}
\caption{
 Example solution for the IGGP \emph{md next} task. This hypothesis states that the value becomes 5 when the player presses the button, and is the true value minus 1 if the player does not act. Magic values are represented in bold.}
\label{fig:md}
\end{figure}

\begin{figure}[t]
\begin{tabular}{l}
\emph{next(A,\textbf{q})$\leftarrow$ my\_true(A,\textbf{q}),does(A,\textbf{robot},\textbf{a})}\\
\emph{next(A,\textbf{p})$\leftarrow$ my\_true(A,\textbf{q}),does(A,\textbf{robot},\textbf{b})}\\
\emph{next(A,\textbf{q})$\leftarrow$ my\_true(A,\textbf{r}),does(A,\textbf{robot},\textbf{c})}\\
\emph{next(A,\textbf{r})$\leftarrow$ my\_true(A,\textbf{r}),does(A,\textbf{robot},\textbf{a})}\\
\emph{next(A,\textbf{r})$\leftarrow$ my\_true(A,\textbf{r}),does(A,\textbf{robot},\textbf{b})}\\
\emph{next(A,\textbf{q})$\leftarrow$ my\_true(A,\textbf{p}),does(A,\textbf{robot},\textbf{b})}\\
\emph{next(A,\textbf{p})$\leftarrow$ my\_true(A,\textbf{p}),does(A,\textbf{robot},\textbf{c})}\\
\emph{next(A,\textbf{r})$\leftarrow$ my\_true(A,\textbf{q}),does(A,\textbf{robot},\textbf{c})}\\
\emph{next(A,B)$\leftarrow$ my\_true(A,C),my\_succ(C,B)}\\
\emph{next(A,\textbf{p})$\leftarrow$ not\_my\_true(A,B),does(A,\textbf{robot},\textbf{a})}
\end{tabular}
\caption{
 Example solution for the IGGP \emph{buttons-next} task. The first clause states that the next value becomes \emph{q} if the current value is \emph{q} and the agent presses the button \emph{a}. Magic values are represented in bold.}
\label{fig:buttons}
\end{figure}

\begin{figure}[t]
\begin{tabular}{l}
\emph{f(A)$\leftarrow$ cell(A,E,\textbf{white},\textbf{rook}),cell(A,B,\textbf{white},\textbf{king}),distance(E,B,\textbf{1})}\
\end{tabular}
\caption{
 Example solution for the \emph{krk} task. This hypothesis describes the concept of rook protected in the chess \emph{krk} endgame. This hypothesis states that the white king protects the white rook when the white king and the white rook are at distance 1 of each other. Magic values are represented in bold.}
\label{fig:krk}
\end{figure}
\paragraph{\emph{KRK}.} The task is to learn a chess pattern in the king-rook-king (\emph{krk}) endgame, which is the chess ending with white having a king and a rook and black having a king. We learn the concept of rook protection by its king \cite{hocquette2020}. An example target solution is presented in Figure \ref{fig:krk}.
This task involves identifying 4 magic values.

\paragraph{Program synthesis: \emph{list}, \emph{powerof2} and \emph{append}.} For \emph{list}, we learn a hypothesis describing the existence of the magic number '7' in a list. 
Figure \ref{fig:intro_target} in the introduction shows an example solution. 
For \emph{powerof2}, we learn a hypothesis which describes whether a number is of the form $2^k$, with k integer. These two problems involve learning a recursive hypothesis. For \emph{append}, we learn that lists must have a particular suffix of size 2.
For \emph{list}, there are 4000 constants in the background knowledge. Examples are lists of size 500. For \emph{powerof2}, examples are numbers between 2 and 1000, there are 1000 constants in the background knowledge. For \emph{append}, examples are lists of size 10, there are 1000 constants in the background knowledge.

\subsubsection{Results}
Table \ref{tab:time} shows the learning times. It shows \name{} can solve each of the tasks in at most 100s, often a few seconds. To put these results into perspective, an approach that precomputes the hypothesis space \cite{aspal} would need to precompute at least $(\text{\#preds} \text{\#constants})^{\text{\#literals}}$ rules. For instance, for \emph{buttons-next}, this approach would need to precompute at least $(5*16)^{10} = O(10^{19})$ rules, which is infeasible. Conversely, \name{} solves this task in 3 seconds.

\popper{} is based on enumeration of possible constant symbols: it uses unary predicate symbols, one for each possible constant symbol. Compared to \popper{}, \name{} has shorter learning times on seven tasks (\emph{md}, \emph{buttons-goal}, \emph{coins-goal}, \emph{gt-centipede-goal}, \emph{gt-centipede-legal}, \emph{gt-centipede-next}, \emph{krk}, \emph{list}, \emph{powerof2} and \emph{append}) and longer learning times on two tasks (\emph{buttons-next} and \emph{coins-next}). A paired t-test confirms the significance of the difference for these ten tasks at the $p<0.01$ level.
For instance, \name{} can solve the \emph{krk} problem in 6s while \popper{} requires almost 35s. 

There are three main reasons for this improvement. 
First, \name{} reasons about magic hypotheses while \popper{} cannot. 
Each magic hypothesis represents the set of its possible instantiations, which alleviates the need to enumerate all possible constant symbols. 
The constraints \name{} formulates eliminate magic hypotheses, which prunes more instantiated programs.
Second, compared to \popper{}, \name{} does not need additional unary predicates to represent constant symbols. 
This feature allows \name{} to learn shorter hypotheses with constant symbols as arguments instead. For instance, in the \emph{krk} experiment, \name{} typically learns a hypothesis with 3 body literals while \popper{} typically needs 6 body literals, including 3 body literals to represent constant symbols. \popper{} thus needs to search up to a larger depth compared to \name. As demonstrated by Proposition \ref{prop:gain}, these two reasons lead to a smaller hypothesis space. Finally, \name{} tests hypotheses against the positive examples and only considers instantiations which, together with the background knowledge, entail at least one positive example. In this sense, \name{} never considers irrelevant constant symbols. For these three reasons, \name{} considers fewer hypotheses which explains the shorter learning times.

However, given a search bound, \name{} searches a larger space than \popper{} since it does not count the magic literals in the program size. \name{} considers the same programs as \popper, but also programs with magic values whose size would exceed the search bound if representing magic values with unary predicate symbols. Therefore, \name{} can require longer running time than \popper{}, which is the case for two tasks (\emph{buttons-next} and \emph{coins-next}).

\ale{} restricts the possible constants to constants appearing in the bottom clause, which is the logically most-specific clause that explains an example. \ale{} also can identify constant symbols through a lazy evaluation procedure \cite{srinivasan1999}, which has inspired our magic evaluation procedure. Therefore, \ale{} does not consider irrelevant constant symbols but only symbols that can be obtained from the examples. Compared to \ale, \name{} has shorter learning times on four tasks (\emph{buttons-next}, \emph{coins-next}, \emph{list}, \emph{append}). A paired t-test confirms the significance of the difference in learning times for these tasks at the $p<0.01$ level. However, in contrast to \ale, \name{} searches for optimal solutions. Moreover, \name{} is given a weaker bias about which variables can be magic variables.


\metagol{} identifies relevant constant symbols by constructing a proof for the positive examples. Therefore, it also considers only relevant constant symbols that can be obtained from the examples. Compared to \name{}, \metagol{} has longer learning times on 6 tasks, similar learning times on 3 tasks, and better learning time on three tasks.

Table \ref{tab:accuracies} shows the predictive accuracies. \name{} achieves higher or equal accuracies than \metagol{}, \ale{}, and \popper{}, apart on \emph{gt-centipede-goal}.
This improvement can be explained by the fact that \name{} can learn in domains other systems cannot handle. For instance, \name{} supports learning with predicate symbols of arity more than two, which is necessary for the IGGP games and the \textit{krk} domain. By contrast, \metagol{} cannot learn hypotheses with arity greater than 2 given the set of metarules provided. Compared to \name, \ale{} struggles to learn recursive hypotheses. However, \ale{} performs well on the tasks which do not require recursion, reaching similar or better accuracy than \name{} on seven tasks (\emph{md}, \emph{buttons-goal}, \emph{gt-centipede-goal}, \emph{gt-centipede-legal}, \emph{gt-centipede-next} \emph{krk}, and \emph{append}).
Finally, compared to \popper{}, \name{} can achieve higher accuracies. For instance, on the list problem, \name{} reaches 100\% accuracy while \popper{} achieves the default accuracy. Since it does not enumerate constant symbols, \name{} can search a smaller space than \popper, and thus its learning time can be shorter. Therefore, it is more likely to find a solution before timeout. Also, according to the Blumer bound \cite{blumer1989}, given two hypotheses spaces of different sizes, searching the smaller space can result in higher predictive accuracy compared to searching the larger one if a target hypothesis is in both.

Given these results, we can positively answer \textbf{Q1} and confirm that \name{} can outperform existing approaches in terms of learning times and predictive accuracies when learning programs with magic values.

\begin{table}[ht]
\centering
\begin{tabular}{l|cccc}
\textbf{Task} & \textbf{\ale} & \textbf{\metagol} & \textbf{\popper} & \textbf{\name}\\
\midrule
\emph{md} & \textbf{0 $\pm$ 0} & timeout & 1 $\pm$ 0 & \textbf{0 $\pm$ 0} \\
\emph{buttons-next} & 32 $\pm$ 1 & timeout & \textbf{3 $\pm$ 0} & 4 $\pm$ 0\\
\emph{coins-next} & timeout & \textbf{0 $\pm$ 0} & \textbf{53 $\pm$ 0} & 99 $\pm$ 1\\
\emph{buttons-goal} & \textbf{0 $\pm$ 0} & \textbf{0 $\pm$ 0} & 1 $\pm$ 0 & \textbf{0 $\pm$ 0}\\
\emph{coins-goal} & \textbf{0 $\pm$ 0} & \textbf{0 $\pm$ 0} & \textbf{0 $\pm$ 0} & \textbf{0 $\pm$ 0}\\
\emph{gt-centipede-goal} & \textbf{0 $\pm$ 0}  & \textbf{0 $\pm$ 0} & 23 $\pm$ 0 & 6 $\pm$ 0\\
\emph{gt-centipede-legal} & \textbf{0 $\pm$ 0} & \textbf{0 $\pm$ 0} & 4 $\pm$ 0 & 1 $\pm$ 0\\
\emph{gt-centipede-next} & \textbf{0 $\pm$ 0} & timeout & 10 $\pm$ 0 & \textbf{0 $\pm$ 0}\\
\emph{krk} & \textbf{0 $\pm$ 0} & 541 $\pm$ 60 & 35 $\pm$ 6 & 6 $\pm$ 0\\
\emph{list} & 66 $\pm$ 1 & 36 $\pm$ 8 &  timeout & \textbf{2 $\pm$ 0}\\
\emph{powerof2} & \textbf{0 $\pm$ 0} & 463 $\pm$ 78 & 18 $\pm$ 0 & \textbf{0 $\pm$ 0}\\
\emph{append} & 1 $\pm$ 0 & \textbf{0 $\pm$ 0} & 298 $\pm$ 49 & \textbf{0 $\pm$ 0}\\
\midrule
\emph{pi} & 4 $\pm$ 1 & \textbf{0 $\pm$ 0} & timeout &  1 $\pm$ 0\\
\emph{equilibrium} & \textbf{0 $\pm$ 0} & \textbf{0 $\pm$ 0} & 209 $\pm$ 7 & 72 $\pm$ 17\\
\emph{drug design} & 5 $\pm$ 1 & timeout &  \textbf{1 $\pm$ 0} & 6 $\pm$ 3\\
\emph{next} & \textbf{0 $\pm$ 0} & timeout & 1 $\pm$ 0 & 25 $\pm$ 0\\
\emph{sumk} & \textbf{0 $\pm$ 0} & timeout & \textbf{0 $\pm$ 0} & 99 $\pm$ 1\\
\end{tabular}
\caption{
Learning times. We round times to the nearest second. The error is standard deviation. Tasks above the horizontal line have finite domains while tasks below the horizontal line have infinite constant domains.}

\label{tab:time}
\end{table}

\begin{table}[ht]
\centering
\begin{tabular}{l|cccc}
\textbf{Task} & \textbf{\ale} & \textbf{\metagol} & \textbf{\popper{}} & \textbf{\name}\\
\midrule
\emph{md} & \textbf{100 $\pm$ 0} & 50 $\pm$ 0 & \textbf{100 $\pm$ 0} & \textbf{100 $\pm$ 0}\\
\emph{buttons-next} & 81 $\pm$ 0 & 50 $\pm$ 0 & \textbf{100 $\pm$ 0} & \textbf{100 $\pm$ 0} \\
\emph{coins-next} & 50 $\pm$ 0 & 50 $\pm$ 0 & \textbf{100 $\pm$ 0} & \textbf{100 $\pm$ 0}\\
\emph{buttons-goal} & \textbf{100 $\pm$ 0} & 50 $\pm$ 0 & 98 $\pm$ 1 & \textbf{100 $\pm$ 0}\\
\emph{coins-goal} & 50 $\pm$ 0 & 50 $\pm$ 0 & \textbf{100 $\pm$ 0} & \textbf{100 $\pm$ 0}\\
\emph{gt-centipede-goal} & \textbf{99 $\pm$ 0} & 50 $\pm$ 0 & 75 $\pm$ 0 & 75 $\pm$ 0\\
\emph{gt-centipede-legal} & \textbf{100 $\pm$ 0} & 50 $\pm$ 0 & \textbf{100 $\pm$ 0}& \textbf{100 $\pm$ 0}\\
\emph{gt-centipede-next} & \textbf{100 $\pm$ 0} & 50 $\pm$ 0 & \textbf{100 $\pm$ 0} & \textbf{100 $\pm$ 0}\\
\emph{krk} & \textbf{100 $\pm$ 0} & 54 $\pm$ 4 & 96 $\pm$ 1 & 99 $\pm$ 0 \\
\emph{list} & 50 $\pm$ 0 & \textbf{100 $\pm$ 0} & 49 $\pm$ 0 & \textbf{100 $\pm$ 0}\\
\emph{powerof2} & 86 $\pm$ 1 & 58 $\pm$ 5 & 84 $\pm$ 1 & \textbf{100 $\pm$ 0}\\
\emph{append} & 95 $\pm$ 1 & \textbf{99 $\pm$ 0} & 96 $\pm$ 1 & 96 $\pm$ 1\\
\midrule
\emph{pi} & \textbf{100 $\pm$ 0} & 50 $\pm$ 0 & 50 $\pm$ 0 & 99 $\pm$ 0\\
\emph{equilibrium} & \textbf{100 $\pm$ 0} & 50 $\pm$ 0 & 62 $\pm$ 1 & 86 $\pm$ 7\\
\emph{drug design} & 63 $\pm$ 7 & 50 $\pm$ 0 & 50 $\pm$ 0 & \textbf{98 $\pm$ 0}\\
\emph{next} & 50 $\pm$ 0 & 50 $\pm$ 0 & 49 $\pm$ 0 & \textbf{100 $\pm$ 0}\\
\emph{sumk} & 50 $\pm$ 0 & 50 $\pm$ 0 & 50 $\pm$ 0 & \textbf{100 $\pm$ 0}\\

\end{tabular}
\caption{
Predictive accuracies. We round to the closest integer. The error is standard deviation.  Tasks above the horizontal line have finite domains while tasks below the horizontal line have infinite constant domains.}

\label{tab:accuracies}
\end{table}
\subsection[\textbf{Q2}: infinite domains]{\textbf{Q2}: learning in infinite domains}
We evaluate the performance of \name{} in infinite domains and compare it against the performance of \popper, \ale{}, and \metagol. This experiment aims to answer \textbf{Q1} and \textbf{Q2}. We consider five tasks. Full descriptions are in the appendix. 
We use a timeout of 600s for each of these tasks.
\subsubsection{Experimental domains}
\paragraph{Learning Pi.}
The goal of this task is to learn a mathematical equation over real numbers expressing the relation between the radius of a disk and its area. 

This task involves identifying the magic value \emph{pi} up to floating-point precision. We allow a precision error of $10^{-3}$. Figure \ref{fig:pi} shows an example solution.


\begin{figure}[t]
\centering
\begin{tabular}{l}
\emph{area(A,B) $\leftarrow$ square(A,C),mult(C,\textbf{3.142},B).}\\
\end{tabular}
\caption{
 Example solution for the \textit{pi} task. The magic constant pi is represented in bold.}
\label{fig:pi}
\end{figure}

\paragraph{Equilibrium.} The task is to identify a relation describing mechanical equilibrium for an object subject to its weight and other forces whose values are known. This task involves identifying the gravitational constant \emph{g} up to floating-point precision. We allow a precision error of $10^{-3}$. Figure \ref{fig:equilibrium} shows an example of the target hypothesis. 

\begin{figure}[t]
\centering
\begin{tabular}{l}
\emph{equilibrium(A) $\leftarrow$ mass(A,B),forces(A,C),sum(C,D),mult(B,\textbf{9.807},D)}\\
\end{tabular}
\caption{
 Example solution for the equilibrium task. The gravitational constant g is represented in bold.}
\label{fig:equilibrium}
\end{figure}
\paragraph{Drug design.} The goal of this task is to identify molecule properties representing suitable medicinal activity. An example is a molecule which is represented by the atoms it contains and the pairwise distance between these atoms. Atoms have varying types. 
Figure \ref{fig:drug} shows an example solution. This task involves identifying two magic values representing the particular atom types ``o'' and ``h'' and one magic value representing a specific distance between two atoms. 

\begin{figure}[t]
\centering
\begin{tabular}{l}
\emph{drug(A) $\leftarrow$ atom(A,B),atom(A,C),atom$\_$type(B,\textbf{o}), atom$\_$type(C,\textbf{h}),}\\
\emph{distance(B,C,\textbf{0.513})}\\
\end{tabular}
\caption{Example solution for the drug design task. Magic values for atom types and an example of magic value for the distance are represented in bold.}
\label{fig:drug}
\end{figure}

\paragraph{Program Synthesis: \emph{next} and \emph{sumk}.}
 For \textit{next}, we learn a hypothesis for identifying the element following a magic value in a list. For example, given the magic value 4.543, we may have the positive example \emph{next([1.246, 4.543, 2.156],2.156)}. Figure \ref{fig:next} shows an example solution. Examples are lists of size 500 of float numbers.
For \textit{sumk}, we learn a relation describing that two elements of a list have a sum equal to $k$, where $k$ is an integer magic value. Examples are lists of size 50 of integer numbers. 
Figure \ref{fig:sumk} shows an example of target hypothesis.

\begin{figure}[t]
\centering
\begin{tabular}{l}
\emph{next(A,B) $\leftarrow$
	head(A,\textbf{4.543}),
	tail(A,C)
head(C,B)}\\
\emph{next(A,B) $\leftarrow$
tail(A,C),next(C,B)}
\end{tabular}
\caption{
 Example solution for the \textit{next\_element} task. An example of magic constant is represented in bold.}
\label{fig:next}
\end{figure}

\begin{figure}[t]
\begin{tabular}{l}
\emph{sumk(A) $\leftarrow$
	member(A,B),
	member(A,C),
	add(B,C,\textbf{612})}
\end{tabular}
\caption{
 Example solution for the \textit{sumk} task. An example of magic constant is represented in bold.}
\label{fig:sumk}
\end{figure}

\subsubsection{Results}
Tables \ref{tab:time} and \ref{tab:accuracies} show the results. They show that, compared to \popper, \name{} achieves higher accuracy\footnote{\name{} does not always achieve maximal accuracy due to floating-point precision errors.}. \popper{} cannot identify hypotheses with magic values in infinite domains because it cannot represent an infinite number of constant symbols. Thus, it achieves the default accuracy. 
\metagol{} cannot learn hypotheses with arity greater than 2 given the metarules provided and therefore struggles on these tasks. It also struggles when the proof length is large, such as when examples are lists of large size. \ale, through the use of lazy evaluation, performs well on the tasks which do not require recursion, especially \emph{pi} and \emph{equilibrium}. However, it struggles on \emph{next} and \emph{sumk} which both require recursion. The learning time of \name{} is better than that's of \ale{} on one of the two tasks \ale{} can solve, but worse on the other. However, in contrast to \ale{}, \name{} searches for optimal hypotheses. Moreover, \name{} searches a larger search space since it is given as bias the types of variables which can be magic variables while \ale{} is given the arguments of some predicate symbols through the mode declarations.

These results demonstrate that \name{} can identify magic values in infinite domains. These results confirm our answer to \textbf{Q1}. Also, we positively answer \textbf{Q2}.


\pgfplotsset{select coords between index/.style 2 args={
    x filter/.code={
        \ifnum\coordindex<#1\def\pgfmathresult{}\fi
        \ifnum\coordindex>#2\def\pgfmathresult{}\fi
    }
}}
\subsection[\textbf{Q3}: scalability]{\textbf{Q3}: scalability with respect to the number of constant symbols} \label{exp2}
We now evaluate how well our approach scales. First, we evaluate how well our approach scales with the number of constant symbols. To do so, we need domains in which we can control the number of constant symbols. 
We consider two domains: \emph{list} and \emph{md}.
In the \emph{list} experiment, described in Section \ref{domains}, we use an increasingly larger set of constant symbols disjoint from $\{7\}$ in the background knowledge. 
In the \emph{md} experiment, also described in Section \ref{domains}, we vary the number of next values available. 
We use a timeout of 60s for each task.
Full details are in the appendix.

\pgfplotstableread{
xs	acc_av	acc_std	time_av	time_std
1	50	0	60.0196621573996	0.004927581200344
100	50	0	60.0284631977018	0.007333017573899
200	50	0	60.0104039716971	0.0011438453605
400	50	0	60.0153843688997	0.00189431293769
800	50	0	60.0156667213989	0.002094642280526
1500	50	0	60.0151284941981	0.003371698981924
3000	50	0	60.033028421698	0.011848317475845
6000	50	0	60.020437543602	0.002682245149408
12000	50	0	60.0168722759001	0.002778042954407
25000	50	0	60.0157730731007	0.0029458450856
50000	50	0	60.0413221393013	0.019301401150295
100000	50	0	60.1375942590006	0.044423887494141
200000	50	0	60.0283714265024	0.003438270892604
400000	50	0	60.0386479754016	0.010180135004445
800000	50	0	60.0444317252011	0.009975289864998
1500000	50	0	60.0648517970025	0.008770669000809
3000000	50	0	60.0460527047981	0.009455747685279
}\reslistaleph
\pgfplotstableread{
xs	acc_av	acc_std	time_av	time_std
1	50.0	0.0	0.3358848748	0.015028766528308200
100	100.0	0.0	1.0171465041	0.11602527273603200
200	100.0	0.0	0.8003416333	0.15109687311224200
400	100.0	0.0	1.0944211875	0.13217524619093400
800	100.0	0.0	1.4247696708	0.1305424251444600
1500	100.0	0.0	1.9392523917	0.05369301316621090
3000	100.0	0.0	2.5640482793	0.1373656687902910
6000	100.0	0.0	3.6738930918	0.038848256713826500
12000	100.0	0.0	4.4081449916	0.33543887253028200
25000	100.0	0.0	4.115231425	0.04600762046823230
50000	100.0	0.0	4.5862747916	0.07211000171498840
100000	100.0	0.0	4.9012878999	0.03673039637401910
200000	100.0	0.0	4.7464308167	0.14172085756024300
400000	100.0	0.0	5.2236428876000000	0.1525404307621810
800000	100.0	0.0	4.84842505	0.10313510590060100
1500000	100.0	0.0	4.665735225100000	0.1652361609426830
3000000	100.0	0.0	4.4492744166	0.07374305575525400
}\reslistmagicpopper

\pgfplotstableread{
xs	acc_av	acc_std	time_av	time_std
1	50.0	0.0	60.08606943949420	0.0016040371636667900
100	100.0	0.0	20.924571299398700	2.2396595516227300
200	100.0	0.0	21.779307434300400	3.585414828805590
400	100.0	0.0	21.288158281499600	3.5269608801199000
800	100.0	0.0	22.99464180650250	3.783334212189010
1500	100.0	0.0	18.41662772450070	3.1922744232395200
3000	100.0	0.0	21.385671431897200	3.1433250941772500
6000	100.0	0.0	26.993440664108500	3.173405217812740
12000	100.0	0.0	14.187998352292900	3.4939868453333800
25000	100.0	0.0	26.94566109459850	2.0316409899829300
50000	100.0	0.0	21.97908923520300	3.1711030576720400
100000	100.0	0.0	14.209471077899800	3.67127216963358
200000	100.0	0.0	19.09530635669940	3.058530383116640
400000	100.0	0.0	19.607858572900300	4.425284965208290
800000	100.0	0.0	23.382287235904400	3.6030603534266100
1500000	100.0	0.0	16.82699034829860	3.223499709541470
3000000	100.0	0.0	22.721790962098700	3.6544577702718500
}\reslistmetagol
\pgfplotstableread{
xs	acc_av	acc_std	time_av	time_std
1	100.0	0.0	0.2914160998	0.00608703031930077
100	100.0	0.0	5.762593091600000	0.4124556589959130
200	100.0	0.0	12.370400995700000	0.9189280169075260
400	100.0	0.0	34.2431743499	0.7142108879754640
800	49.905	0.10965856099730700	60.0	0.0
1500	49.96	0.05567764362830030	60.0	0.0
3000	50.0	0.0	60.0	0.0
6000	50.005	0.004999999999999720	60.0	0.0
12000	50.0	0.0	60.0	0.0
25000	50.005	0.004999999999999720	60.0	0.0
50000	50.0	0.0	60.0	0.0
100000	50.0	0.0	60.0	0.0
200000	50.0	0.0	60.0	0.0
400000	50.0	0.0	60.0	0.0
800000	50.0	0.0	60.0	0.0
1500000	50.0	0.0	60.0	0.0
3000000	50.0	0.0	60.0	0.0
}\reslistpopper

\pgfplotstableread{
xs	acc_av	acc_std	time_av	time_std
1	100.0	0.0	0.2595135940995530	0.02555228986512030
100	74.9	0.4131854574186050	3.560789095592920	0.34038431209317200
200	75.055	0.23015091667087600	5.611188465188020	0.4767991578242950
400	75.02	0.24924329925240200	12.839529921807100	1.5972492451592900
800	75.075	0.22867371223353800	45.91043485819830	1.7551938362255500
1500	52.515	2.5150000000000000	58.32509962669570	1.7475099072784200
3000	50.0	0.0	60.06670204520340	0.0052235813744339100
6000	50.0	0.0	60.09591264319610	0.008730716213903960
12000	50.0	0.0	60.165004598110700	0.010733638991781000
25000	50.0	0.0	60.180206411291100	0.01477998539548870
50000	50.0	0.0	60.18937632940940	0.010240955422180000
100000	50.0	0.0	60.183367063506700	0.01597956617415490
200000	50.0	0.0	60.20261552939770	0.01677784896252140
400000	50.0	0.0	60.20394974201340	0.015318398932069300
800000	50.0	0.0	60.633265106607	0.07082182558826240
1500000	50.0	0.0	60.58990817329610	0.07463141623735580
3000000	50.0	0.0	60.77924991759360	0.11664080259119400
}\resmdaleph
\pgfplotstableread{
xs	acc_av	acc_std	time_av	time_std
1	50	0.0	60.06819796069760	0.00448850723562413
100	50	0.0	60.07509338720700	0.003532982947365050
200	50	0.0	60.068456090200900	0.0036970833298282600
400	50	0.0	60.08047119041440	0.00544937086968727
800	50	0.0	60.08385397831440	0.010241619008702000
1500	50	0.0	60.09353470639330	0.010738651305679600
3000	50	0.0	60.13745172658820	0.005532835774287810
6000	50	0.0	60.13762015990800	0.009747608724157390
12000	50	0.0	60.147577193903300	0.008713233453232360
25000	50	0.0	60.156056610000100	0.012819587299430400
50000	50	0.0	60.18541577400760	0.01172271059290300
100000	50	0.0	60.2147604470083	0.015708363750114200
200000	50	0.0	60.18437499210120	0.012163786476151700
400000	50	0.0	60.272693820600400	0.014800830048185800
800000	50	0.0	60.42626096989840	0.07009202726529650
1500000	50	0.0	60.76734465589400	0.06761751034344250
3000000	50	0.0	60.97252165870160	0.08146098969178640
}\resmdmetagol
\pgfplotstableread{
xs	acc_av	acc_std	time_av	time_std
1	100.0	0.0	0.4118878125	0.011449841845835200
100	100.0	0.0	0.8087709749	0.11022156655150100
200	97.55485294117650	2.445147058823530	0.8550718838	0.028131957779192800
400	100.0	0.0	1.5139596373	0.05836325074923590
800	100.0	0.0	3.2006476211	0.18478403371663900
1500	100.0	0.0	4.7429955709	0.5321205855165620
3000	99.99919972550590	0.0008002744941506990	7.3149174376	1.1576570573322100
6000	99.9971395201992	0.0006340871498292830	7.828607325	1.2880141363886100
12000	99.99876990048690	0.0003435545362201480	22.5610895999	6.5692234977604400
25000	99.99903996085840	8.02112956448495E-05	19.140300295700000	4.275267564901540
50000	99.99946998942980	6.4206055330287E-05	51.6651975292	3.2797159089627000
100000	97.54973499772500	2.4500816674501700	45.968405591600000	7.470000624361240
200000	99.999874999445	3.59398614202979E-05	54.6112144958	5.3887855042
400000	99.99996999994	1.3333359999271E-05	60.0	0.0
800000	99.99997999999	8.16496989210917E-06	60.0	0.0
1500000	99.999979999985	1.10554204911877E-05	60.0	0.0
3000000	99.999995	5.00000000016598E-06	60.0	0.0
}\resmdmagicpopper
\pgfplotstableread{
xs	acc_av	acc_std	time_av	time_std
1	100.0	0.0	2.2068230124	0.07005065878013830
100	76.04355769230770	0.2624418707749040	60.0	0.0
200	51.0057734204793	0.04985550092839760	60.0	0.0
400	50.354084158415800	0.06376141053785630	60.0	0.0
800	50.07494402985080	0.03182046397783070	60.0	0.0
1500	50.05996497669950	0.025596658564794200	60.0	0.0
3000	54.90835944704360	3.272451592758270	60.0	0.0
6000	55.124194869033700	3.416587819989880	60.0	0.0
12000	70.12333985589740	3.360160872978790	60.0	0.0
25000	50.0	0.0	60.0	0.0
50000	50.0	0.0	60.0	0.0
100000	50.0	0.0	60.0	0.0
200000	50.0	0.0	60.0	0.0
400000	50.0	0.0	60.0	0.0
800000	50.0	0.0	60.0	0.0
1500000	50.0	0.0	60.0	0.0
3000000	50.0	0.0	60.0	0.0
}\resmdpopper

\pgfplotsset{scaled x ticks=false}

\begin{figure}
\begin{tikzpicture}
\begin{customlegend}[legend columns=5,legend style={nodes={scale=1, transform shape},align=left,column sep=0ex},
        legend entries={\popper, \name, \ale, \metagol}]
        \addlegendimage{black,mark=triangle*}
        \addlegendimage{red,mark=diamond*}
        \addlegendimage{blue,mark=square*}
        \addlegendimage{green,mark=otimes*}
        \end{customlegend}
\end{tikzpicture}\\
\smallbreak
  \begin{minipage}{0.45\textwidth}
\resizebox{\columnwidth}{!}{
\begin{tikzpicture}
\begin{axis}[
  legend style={at={(0.5,0.35)},anchor=west},
   legend style={font=\small},
  xtick={10,100,1000,10000,100000,1000000,10000000},
  ytick={1,10,100},
  xlabel=Number of constant symbols,
  ylabel=Learning time (s),
    ymode=log,
      xmode=log,
  ]
\addplot+[select coords between index={1}{16},
blue,mark=square*,
                error bars/.cd,
                y dir=both,
                error mark,
                y explicit]table[x=xs,y=time_av,y error=time_std] {\reslistaleph};
    \addplot[select coords between index={1}{16},
    green,mark=otimes*,
                error bars/.cd,
                y dir=both,
                error mark,
                y explicit]table[x=xs,y=time_av,
    y error=time_std] {\reslistmetagol};
    \addplot[select coords between index={1}{16},
    black,mark=triangle*,
                error bars/.cd,
                y dir=both,
                error mark,
                y explicit]table[x=xs,y=time_av,
    y error=time_std] {\reslistpopper};
    \addplot[select coords between index={1}{16},
    red,mark=diamond*,
                error bars/.cd,
                y dir=both,
                error mark,
                y explicit]table[x=xs, y=time_av,
    y error=time_std] {\reslistmagicpopper};
  
\end{axis}
\end{tikzpicture}}
\caption{List: learning time versus the number of constant symbols. Axes are log scaled.}
\label{fig:list_time}
\end{minipage}\hfill
  \begin{minipage}{0.45\textwidth}
\resizebox{\columnwidth}{!}{
\begin{tikzpicture}
\begin{axis}[
  legend style={at={(0.5,0.5)},anchor=west},
   legend style={font=\small},
  xtick={10,100,1000,10000,100000,1000000,10000000},
  xlabel=Number of constant symbols,
  ylabel=Accuracy (\%),
  xmode=log
  ]
  \addplot+[select coords between index={1}{16},
  blue,mark=square*,
                error bars/.cd,
                y dir=both,
                error mark,
                y explicit]table[x=xs,y=acc_av,
  y error=acc_std] {\reslistaleph};
    \addplot[select coords between index={1}{16},
    green,mark=otimes*,
                error bars/.cd,
                y dir=both,
                error mark,
                y explicit]table[x=xs,y=acc_av,
    y error=acc_std] {\reslistmetagol};
    \addplot[select coords between index={1}{16},
    black,mark=triangle*,
                error bars/.cd,
                y dir=both,
                error mark,
                y explicit]table[x=xs,y=acc_av,
    y error=acc_std] {\reslistpopper};
    \addplot[select coords between index={1}{16},
    red,mark=diamond*,
                error bars/.cd,
                y dir=both,
                error mark,
                y explicit]table[x=xs
,y=acc_av,
    y error=acc_std] {\reslistmagicpopper};
\end{axis}
\end{tikzpicture}}
\caption{List: accuracy versus the number of constant symbols. The horizontal axis is log scaled.}
\label{fig:list_acc}
  \end{minipage}\\
\smallbreak
  \begin{minipage}{0.45\textwidth}
\resizebox{\columnwidth}{!}{
\begin{tikzpicture}
\begin{axis}[
  legend style={at={(0.5,0.2)},anchor=west},
   legend style={font=\small},
  xtick={10,100,1000,10000,100000,1000000,10000000},
  xlabel=Number of constant symbols,
  ylabel=Learning time (s),
    ymode=log,
      xmode=log
  ]
\addplot+[select coords between index={1}{16},
blue,mark=square*,
                error bars/.cd,
                y dir=both,
                error mark,
                y explicit]table[x=xs,y=time_av,y error=time_std] {\resmdaleph};
    \addplot[select coords between index={1}{16},
    green,mark=otimes*,
                error bars/.cd,
                y dir=both,
                error mark,
                y explicit]table[x=xs,y=time_av,
    y error=time_std] {\resmdmetagol};
    \addplot[select coords between index={1}{16},
    black,mark=triangle*,
                error bars/.cd,
                y dir=both,
                error mark,
                y explicit]table[x=xs,y=time_av,
    y error=time_std] {\resmdpopper};
    \addplot[select coords between index={1}{16},
    red,mark=diamond*,
                error bars/.cd,
                y dir=both,
                error mark,
                y explicit]table[x=xs, y=time_av,
    y error=time_std] {\resmdmagicpopper};
\end{axis}
\end{tikzpicture}}
\caption{Md: learning time versus the number of constant symbols. Axes are log scaled.}
\label{fig:md_time}
\end{minipage} \hfill
  \begin{minipage}{0.45\textwidth}
\resizebox{\columnwidth}{!}{
\begin{tikzpicture}
\begin{axis}[
  legend style={at={(0.5,0.26)},anchor=west},
   legend style={font=\small},
  xtick={10,100,1000,10000,100000,1000000,10000000},
      xmode=log,
  xlabel=Number of constant symbols,
  ylabel=Accuracy (\%),
  ]
  \addplot+[select coords between index={1}{16},
  blue,mark=square*,
                error bars/.cd,
                y dir=both,
                error mark,
                y explicit]table[x=xs,y=acc_av,
  y error=acc_std]{\resmdaleph};
    \addplot[select coords between index={1}{16},
    green,mark=otimes*,
                error bars/.cd,
                y dir=both,
                error mark,
                y explicit]table[x=xs,y=acc_av,
    y error=acc_std] {\resmdmetagol};
    \addplot[select coords between index={1}{16},
    black,mark=triangle*,
                error bars/.cd,
                y dir=both,
                error mark,
                y explicit]table[x=xs,y=acc_av,
    y error=acc_std] {\resmdpopper};
    \addplot[select coords between index={1}{16},
    red,mark=diamond*,
                error bars/.cd,
                y dir=both,
                error mark,
                y explicit]table[x=xs,y=acc_av,
    y error=acc_std] {\resmdmagicpopper};
\end{axis}
\end{tikzpicture}}
\caption{Md: accuracy versus the number of constant symbols. The horizontal axis is log scaled.}
\label{fig:md_accuracy}
  \end{minipage}
\end{figure}

\subsubsection{Results}
Figures \ref{fig:list_time} and \ref{fig:md_time} show the learning times of \popper{}, \name{}, \ale{}, and \metagol{} versus the number of constant symbols. 
These results show that \name{} has a significantly shorter learning time than \popper. 
\popper{} needs a unary predicate symbol in the background knowledge for each constant symbol, thus the search space grows with the number of constant symbols. 
Moreover, \popper{} considers individually and exhaustively each of the candidate constant symbols. Therefore, \popper{} cannot scale to large background knowledge including a large number of constant symbols. 
It is overwhelmed by 800 constant symbols in the \textit{list} domain and 200 constant symbols in the \textit{md} domain, and it systematically reaches timeout after. 
By contrast, \name{} does not consider every constant symbol but only relevant ones which can be identified from executing the hypotheses on the examples. Thus, it can scale better and can learn from domains with more than 3 million constant symbols. 
This result supports Proposition \ref{prop:gain}, which demonstrated that allowing magic variables can reduce the size of the hypothesis space compared to adding unary predicate symbols and that the difference in the size of the search spaces increases with the number of constant symbols available in the background knowledge.

Figures \ref{fig:list_acc} and \ref{fig:md_accuracy} show the predictive accuracy of \popper{}, \name{}, \ale{}, and \metagol{} versus the number of constant symbols. \popper{} rapidly converges to the default accuracy (50\%) since it reaches timeout. Conversely, \name{} constantly achieves maximal accuracy and outperforms all other systems. In the \textit{md} domain, negative examples must be sampled from a large number of constant symbols, which also can explain the drops in accuracy. \ale{} struggles to learn recursive programs which explains its low predictive accuracy in the list domain. Moreover, \ale{} is based on the construction of a bottom clause. The bottom clause can grow very large in both domains when the number of constant symbols augments, which can overwhelm the search. 
\metagol{} can learn programs with constant symbols using the \textit{curry} metarules. It performs well and scales to a large number of constant symbols in the list experiment. However, the metarules provided are not expressive enough to support learning with higher-arity predicates, which in particular prevents \metagol{} from learning a solution for \textit{md} for any of the numbers of constants tested.

These results confirm our answer to \textbf{Q1}. They also show that the answer to \textbf{Q3} is that \name{} can scale well with the number of constant symbols, up to millions of constant symbols.

\subsection[\textbf{Q3}: scalability]{\textbf{Q3}: scalability with respect to the number of magic values}
To evaluate scalability with respect to the number of magic values, we vary the number of magic values within the target hypothesis.
We vary the number of magic values along two dimensions (i) the number of magic values within one clause, and (ii) the number of magic values in different independent clauses. 


\subsubsection{Magic values in one clause}\label{oneclause}
 We first evaluate scalability with respect to the number of magic values in the same clause. We learn hypotheses of the form presented in Figure \ref{fig:target_hyp1}, where the number of body literals varies. There are 100 constants in the background knowledge. Lists have size 100. We use a timeout of 60s for each task. Full experimental details are in the appendix.
    

\paragraph{Results}
Figures \ref{fig:q3b_time} and \ref{fig:q3b_acc} show the learning times and predictive accuracies.
These results show that, for a small number of magic values, \name{} achieves shorter learning times than \popper. This results in higher predictive accuracies since \popper{} might not find a solution before timeout. From 3 magic values, both systems reach timeout and their performance is similar. When increasing the number of magic values, the number of body literals increases and more search is needed. In particular, \popper{} requires twice as many body literals compared to \name{}, as it needs unary predicates to represent constant symbols. \name{} evaluates magic values within the same clause jointly. For each positive example, it considers the cartesian product of their possible values. The complexity is of the order $O(n^k)$, where $n$ is the size of lists and $k$ is the number of magic values. The complexity is exponential in the number of magic values, which limits scalability when increasing the number of magic values. These results show that \name{} can scale as well as \popper{} with respect to the number of magic values in the same clause, thus answering \textbf{Q3}. However, scalability is limited for both systems. More generally, scalability with respect to the number of magic values is limited for large inseparable programs, such as programs with several magic values in the same clause or in recursive clauses with the same head predicate symbol.
\begin{figure}[t]
\begin{tabular}{l}
\emph{ f(A) $\leftarrow$ member(A,\textbf{1}),member(A,\textbf{2}),member(A,\textbf{3})}\\
\end{tabular}
\caption{
 Example solution. Examples of magic values are represented in bold.}
\label{fig:target_hyp1}
\end{figure}

\pgfplotstableread{
xs	acc_av	acc_std	time_av	time_std
1	100.0	0.0	0.2997728126	0.007883544317578260
2	100.0	0.0	3.0303366333	0.08790435314730360
3	50.0	0.0	60.0	0.0
4	50.0	0.0	60.0	0.0
5	50.0	0.0	60.0	0.0
}\listmagicvaluesmagicpopper
\pgfplotstableread{
xs	acc_av	acc_std	time_av	time_std
1	100.0	0.0	0.3220847124	0.006697456444358020
2	70.0	8.164965809277260	53.4873884834	3.0096609578278200
3	50.0	0.0	60.0	0.0
4	50.0	0.0	60.0	0.0
5	50.0	0.0	60.0	0.0
}\reslistmagicvaluespopper

\begin{figure}
  \begin{minipage}{0.45\textwidth}
\resizebox{\columnwidth}{!}{
\begin{tikzpicture}
\begin{axis}[
  legend style={at={(0.5,0.2)},anchor=west},
  legend style={font=\small},
xtick={1,2,3,4,5},
  xlabel=Number of magic values,
  ylabel=Learning time (s),
    log ticks with fixed point
  ]
    \addplot[select coords between index={0}{3},
    red,mark=diamond*,
                error bars/.cd,
                y dir=both,
                error mark,
                y explicit]table[x=xs,y=time_av,
    y error=time_std] {\listmagicvaluesmagicpopper};
  \addlegendentry{\name}
      \addplot[select coords between index={0}{3},
      black,mark=triangle*,
                error bars/.cd,
                y dir=both,
                error mark,
                y explicit]table[x=xs,y=time_av,
    y error=time_std] {\reslistmagicvaluespopper};
      \addlegendentry{\popper}
\end{axis}
\end{tikzpicture}}
\caption{Same clause: learning time versus the number of magic values.}
\label{fig:q3b_time}
\end{minipage}\hfill
  \begin{minipage}{0.45\textwidth}
\resizebox{\columnwidth}{!}{
\begin{tikzpicture}
\begin{axis}[
  legend style={at={(0.5,0.8)},anchor=west},
  legend style={font=\small},
xtick={1,2,3,4,5},
  xlabel=Number of magic values,
  ylabel=Accuracy
  ]
    \addplot[select coords between index={0}{3},
    red,mark=diamond*,
                error bars/.cd,
                y dir=both,
                error mark,
                y explicit]table[x=xs,y=acc_av,
    y error=acc_std] {\listmagicvaluesmagicpopper};
  \addlegendentry{\name}
      \addplot[select coords between index={0}{3},
      black,mark=triangle*,
                error bars/.cd,
                y dir=both,
                error mark,
                y explicit]table[x=xs,y=acc_av,
    y error=acc_std] {\reslistmagicvaluespopper};
      \addlegendentry{\popper}
\end{axis}
\end{tikzpicture}}
\caption{Same clause: accuracy versus the number of magic values.}
\label{fig:q3b_acc}
  \end{minipage}\hfill
\end{figure}

\subsubsection{Magic values in multiple clauses}
  We now evaluate scalability with respect to the number of magic values in different independent clauses. We learn hypotheses of the form presented in Figure \ref{fig:target_hyp2}, where the number of clauses varies. There are 500 constants in the background knowledge. Lists have size 500.  Each clause is independent. 
  We use a timeout of 60s for each task. Full experimental details are in the appendix.
 
\begin{figure}[t]
\begin{tabular}{l}
\emph{ f(A) $\leftarrow$ head(A,\textbf{1}).}\\ 
\emph{ f(A) $\leftarrow$ head(A,\textbf{2}).}\\ 
\emph{ f(A) $\leftarrow$ head(A,\textbf{3}).}\\ 
\emph{ f(A) $\leftarrow$ head(A,\textbf{4}).}\\ 
\emph{ f(A) $\leftarrow$ head(A,\textbf{5}).}\\ 
\emph{ f(A) $\leftarrow$ head(A,\textbf{6}).}\\ 
\end{tabular}
\caption{
 Example solution. Magic values are represented in bold.}
\label{fig:target_hyp2}
\end{figure}
\paragraph{Results}
Figure \ref{fig:list_magicvalue_time} shows the learning times. The accuracy is maximal for both systems for any of the numbers of magic values tested.
This result shows that \name{} and \popper{} both can handle a large number of magic values in different clauses, up to at least 70. 
Moreover, \name{} significantly outperforms \popper{} in terms of learning times. For instance, \name{} can learn a hypothesis with 50 magic values in 50 different clauses in about 2s, while \popper{} requires 14s.
 This result shows that \name{} can scale well, in particular better than \popper, with respect to the number of magic values in different clauses, thus answering \textbf{Q3}.
As the number of magic values increases, the target hypothesis has more clauses. Both systems must consider an increasingly larger number of programs to test. However, \name{} considers magic programs and only considers instantiations which cover at least one example, which is more efficient than enumerating all possible instantiations. 

We use a version of \popper{} \cite{popper+} which learns non-separable programs independently and then combines them. 
This strategy is efficient to learn disjunctions of independent clauses, which explains the difference in scale from the previous experiment.
For non-separable hypotheses, \name{} must evaluate magic variables jointly as described in the previous experiment.
\pgfplotstableread{
xs	acc_av	acc_std	time_av	time_std
5	100.0	0.0	1.7619632168	0.018611705721127900
10	100.0	0.0	1.8790985166	0.02455645038886150
15	100.0	0.0	1.9890986249	0.013799543023280300
20	100.0	0.0	1.9827193460000000	0.01341862276101280
25	100.0	0.0	1.9904331539	0.012960566406395300
30	100.0	0.0	2.0263757331	0.020768590155135700
35	100.0	0.0	1.9876477707	0.011404918983647200
40	100.0	0.0	2.021475075	0.013561998627379300
45	100.0	0.0	2.0686245624	0.01173977313745870
50	100.0	0.0	2.1170542916000000	0.03169658680182150
55	100.0	0.0	2.1132909959	0.014206934369398600
60	100.0	0.0	2.1360024459	0.026081126377417600
65	100.0	0.0	2.1066406331	0.012270262739178600
70	100.0	0.0	2.1539160125	0.01691760956715420
75	99.475	0.3641771546926020	2.1613781666	0.030936395583298600
80	99.275	0.4002256308078670	2.1707155667000000	0.013109723497410800
85	99.53	0.2459900630332690	2.2008643501	0.023820250821059700
90	99.78	0.14836142056178500	2.2211090416	0.03409395484441000
95	99.205	0.2378666760089690	2.5410250792	0.08759250057685380
100	99.45	0.26119384032213	2.2564767957	0.0238421200697588
}\listmagicvaluesmagicpopper
\pgfplotstableread{
xs	acc_av	acc_std	time_av	time_std
5	100.0	0.0	11.275889833100000	0.06621272757957430
10	100.0	0.0	12.0189758416	0.08792549583505490
15	100.0	0.0	12.533483383400000	0.08984462843430740
20	100.0	0.0	12.514098525	0.044062268208517000
25	100.0	0.0	12.7023848625	0.06035639343944980
30	100.0	0.0	12.7791507711	0.07319893314842420
35	100.0	0.0	12.8378035416	0.04465395289805530
40	100.0	0.0	12.9737383584	0.056179543766528800
45	100.0	0.0	13.2042921208	0.06659520772998030
50	100.0	0.0	13.7463091295	0.20914073991178600
55	100.0	0.0	13.5214495	0.09596112743740550
60	100.0	0.0	13.6713006499	0.15191620314140500
65	100.0	0.0	13.774732604	0.07067388412736670
70	100.0	0.0	13.8707316752	0.07622371505022780
75	100.0	0.0	14.0601705001	0.09058326398266190
80	100.0	0.0	14.2778826666	0.05920382714265410
85	100.0	0.0	14.4130523875	0.054152048225783800
90	100.0	0.0	15.2992172794	0.4309865422514030
95	100.0	0.0	18.750745571	1.4725529017377100
100	100.0	0.0	14.8592286415	0.20967156010177200
}\reslistmagicvaluespopper

\pgfplotstableread{
xs	acc_av	acc_std	time_av	time_std
10	100.0	0.0	0.1445624208	0.002594074213159980
1000	100.0	0.0	14.655078591800000	0.5305318683624830
2000	100.0	0.0	38.8040669416	2.6410420090066000
3000	76.39	7.870220948472650	57.5010817875	0.9198756742686690
4000	53.11	0.14977761292440700	60.0	0.0
5000	52.980000000000000	0.08137703743822460	60.0	0.0
6000	52.845	0.14840822079655800	60.0	0.0
7000	53.105	0.05449260908090640	60.0	0.0
8000	53.06	0.1723691129846390	60.0	0.0
9000	52.92	0.13666666666666700	60.0	0.0
10000	52.92	0.1150362261782490	60.0	0.0
}\reslistexmagicpopper
\pgfplotstableread{
xs	acc_av	acc_std	time_av	time_std
10	100.0	0.0	2.3302874627000000	0.007125860906923720
1000	100.0	0.0	8.8535565208	0.11406689210851900
2000	100.0	0.0	15.0194121875	0.09490595266699360
3000	100.0	0.0	21.4507058458	0.20155729151845400
4000	100.0	0.0	27.6745439789	0.3313083060390910
5000	100.0	0.0	35.3407251873	0.5162363129250320
6000	100.0	0.0	44.828193725	0.28086273988321100
7000	95.2	4.8	50.9588159041	1.8909150918549600
8000	95.25	4.75	54.6026381374	0.9891400793479330
9000	56.61	4.821876075646170	59.9367865875	0.06321341250000000
10000	56.75	4.806442435823910	59.4805389833	0.5194610167
}\reslistexpopper

\begin{figure}
\begin{minipage}{0.45\textwidth}
\resizebox{\columnwidth}{!}{
\begin{tikzpicture}
\begin{axis}[
xtick={10,20,30,40,50,60,70,80,90,100},
ytick={0,5,10,15,20},
  xlabel=Number of magic values,
  ylabel=Learning time (s),
    log ticks with fixed point
  ]
    \addplot[select coords between index={0}{17},
    red,mark=diamond*,
                error bars/.cd,
                y dir=both,
                error mark,
                y explicit]table[x=xs,y=time_av,
    y error=time_std] {\listmagicvaluesmagicpopper};
      \addplot[select coords between index={0}{17},
      black,mark=triangle*,
                error bars/.cd,
                y dir=both,
                error mark,
                y explicit]table[x=xs,y=time_av,
    y error=time_std] {\reslistmagicvaluespopper};
\end{axis}
\end{tikzpicture}}
\caption{Multiple clauses: learning time versus the number of magic values.}
\label{fig:list_magicvalue_time}
\end{minipage}\hfill
  \begin{minipage}{0.45\textwidth}
\begin{tikzpicture}
\begin{customlegend}[legend columns=2,legend style={nodes={scale=1, transform shape},align=right,column sep=0ex},
        legend entries={\popper, \name}]
        \addlegendimage{black,mark=triangle*}
        \addlegendimage{red,mark=diamond*}
        \end{customlegend}
\end{tikzpicture}
\end{minipage}\hfill\\

\vspace{0.2cm}
  \begin{minipage}{0.45\textwidth}
\resizebox{\columnwidth}{!}{
\begin{tikzpicture}
\begin{axis}[
xtick={0,2000,4000,6000,8000,10000},
  xlabel=Number of examples,
  ylabel=Learning time (s),
  ytick={10,20,30,40,50,60},
    log ticks with fixed point
  ]
    \addplot[
    black,mark=triangle*,
                error bars/.cd,
                y dir=both,
                error mark,
                y explicit]table[x=xs,y=time_av,
    y error=time_std] {\reslistexpopper};
    \addplot[
    red,mark=diamond*,
                error bars/.cd,
                y dir=both,
                error mark,
                y explicit]table[x=xs, y=time_av,
    y error=time_std] {\reslistexmagicpopper};
\end{axis}
\end{tikzpicture}}
\caption{Learning time versus the number of examples}
\label{fig:list_ex_time}
\end{minipage}\hfill
 \begin{minipage}{0.45\textwidth}
\resizebox{\columnwidth}{!}{
\begin{tikzpicture}
\begin{axis}[
xtick={0,2000,4000,6000,8000,10000},
  ytick={50,60,70,80,90,100},
  xlabel=Number of examples,
  ylabel=Accuracy
  ]
    \addplot[
    black,mark=triangle*,
                error bars/.cd,
                y dir=both,
                error mark,
                y explicit]table[x=xs,y=acc_av,
    y error=acc_std] {\reslistexpopper};
    \addplot[
    red,mark=diamond*,
                error bars/.cd,
                y dir=both,
                error mark,
                y explicit]table[x=xs,y=acc_av,
    y error=acc_std] {\reslistexmagicpopper};
\end{axis}
\end{tikzpicture}}
\caption{Accuracy versus the number of examples}
\label{fig:list_ex_acc}
  \end{minipage}\hfill
\end{figure}

\subsection[\textbf{Q3}: scalability]{\textbf{Q3}: scalability with respect to the number of examples}\label{examples}
This experiment aims to evaluate how well \name{} scales with the number of examples. We learn the same hypothesis as in Section \ref{exp2}. This task involves learning a recursive hypothesis to identify a magic value in a list. We compare \popper{} and \name.
We use the same material and methods as in Section \ref{exp2}.
We vary the number of examples: for $n$ between 1 and 3000, we sample $n$ positive examples and $n$ negative ones.
Lists have size at most 50, and there are 200 constant symbols in the background knowledge. We use a timeout of 60s for each task.

\subsubsection{Results}
Figures \ref{fig:list_ex_time} and \ref{fig:list_ex_acc} show the results.
They show both \name{} and \popper{} can learn with up to thousands of examples. However, \name{} reaches timeout from 4000 examples while \popper{} reaches timeout from 9000 examples. Their accuracy consequently drops to the default accuracy from these points respectively. This result shows that \name{} has worse scalability than \popper{} with respect to the number of examples, thus answering \textbf{Q3}. 
For both \popper{} and \name{}, we observe a linear increase in the learning time with the number of examples. When increasing the number of examples, executing the candidate hypotheses over the examples takes more time. In particular, \name{} searches for substitutions for the magic variables which cover at least one positive example. Therefore potentially more bindings for magic variables can be identified. Then, more bindings are tried out over the remaining examples as the number of examples increases. \name{} eventually needs to consider every constant symbol as a candidate constant. Moreover, since \name{} does not take in account the magic literals into the program size, it can consider a larger number of programs with constant symbols than \popper{} for any given program size bound, which also explains how its learning time increases faster than the learning time of \popper. This result highlights one limitation of \name. 

\subsection[\textbf{Q4: bias}]{\textbf{Q4}: effect of the bias about magic variables} \label{bias}
In contrast to mode-directed approaches \cite{muggleton1995,srinivasan1999,aspal}, \name{} does not need to be provided as input which variables should be magic variables but instead can automatically identify them. 
It can, however, use this additional information if given as input. 
We investigate the impact of this additional bias on learning performance and thus aim to answer \textbf{Q4}.
\subsubsection{Material and Methods}
We consider the domains presented in Section \ref{domains}.
We compare three variants of \name{}:
\begin{description}
    \item[\textbf{All}:] we allow any variable to potentially be a magic variable.
    \item[\textbf{Types}:] we allow any variable of types manually chosen to potentially be a magic variable. For instance, for \textit{md}, we allow any variable of type \textit{agent}, \textit{action} and \textit{int} to potentially be a magic variable.
    \item[\textbf{Arguments}:] we manually specify a list of arguments of some predicates symbols that can potentially be magic variables. For instance, for \textit{md}, we flag the second argument of \emph{next} and the second and third arguments of \emph{does}. 
\end{description}
\textit{Arguments} is most closely related to mode declarations approaches, which expect a specification for each argument of each predicate. 
However, the specifications of \textit{Arguments} are more flexible since \name{} considers the flagged variables as potential magic variables and searches for a subset of these variables to bind to constant symbols. 
By contrast, mode declarations are stricter and specify exactly which arguments must be constants. \textit{Types} is comparable to \popper{}, which is provided with types for the unary predicates in our experiments. \textit{Types}, \textit{Arguments} and mode declarations require a user to specify some information about which variables can be bound to constant symbols.

The variables which may be a magic variable in \textit{Arguments} are a subset of those of \textit{Types}, which themselves are a subset of those of \textit{All}. In this sense, the search space is increasingly larger. We compare learning times and predictive accuracies for each of these systems. We provide learning times of \popper{} as a baseline. We use a timeout of 600s per task.

\begin{table}[ht]
\centering
\begin{tabular}{l|cccc}
\textbf{Task} & \textbf{\textit{All}} & \textbf{ \textit{Types}} & \textbf{ \textit{Arguments}}  & \textbf{\popper}\\
\midrule
\emph{md} & \textbf{0 $\pm$ 0} & \textbf{0 $\pm$ 0}  & \textbf{0 $\pm$ 0} & 1 $\pm$ 0\\
\emph{buttons-next} &  6 $\pm$ 0 & 4 $\pm$ 0 & \textbf{2 $\pm$ 0} & 3 $\pm$ 0\\
\emph{coins-next} & timeout & 139 $\pm$ 11 & 97 $\pm$ 5 & \textbf{80 $\pm$ 12}\\
\emph{buttons-goal} & \textbf{0 $\pm$ 0} & \textbf{0 $\pm$ 0} & \textbf{0 $\pm$ 0} & 1 $\pm$ 0\\
\emph{coins-goal} & \textbf{0 $\pm$ 0} & \textbf{0 $\pm$ 0} & \textbf{0 $\pm$ 0} & \textbf{0 $\pm$ 0}\\
\emph{gt-centipede-goal} & 8 $\pm$ 0 & 6 $\pm$ 0 & \textbf{3 $\pm$ 0} & 23 $\pm$ 0\\
\emph{gt-centipede-legal} & 2 $\pm$ 0 & 1 $\pm$ 0 & \textbf{0 $\pm$ 0} & 4 $\pm$ 0\\
\emph{gt-centipede-next} & \textbf{0 $\pm$ 0} & \textbf{0 $\pm$ 0} & \textbf{0 $\pm$ 0} & 10 $\pm$ 0\\
\emph{krk} & 30 $\pm$ 2 & 8 $\pm$ 1  & \textbf{7 $\pm$ 0} & 40 $\pm$ 6\\
\emph{list} & timeout & 2 $\pm$ 0 & \textbf{1 $\pm$ 0} & timeout\\
\emph{powerof2} & \textbf{0 $\pm$ 0} & \textbf{0 $\pm$ 0} & \textbf{0 $\pm$ 0} & 18 $\pm$ 0 \\
\emph{append} & \textbf{0 $\pm$ 0} & \textbf{0 $\pm$ 0} & \textbf{0 $\pm$ 0} & 262 $\pm$ 43\\
\end{tabular}
\caption{
Learning times. We round times to the nearest second. The error is standard deviation.}
\label{tab:time_bias}
\end{table}

\begin{table}[ht]
\centering
\begin{tabular}{l|cccc}
\textbf{Task} & \textbf{\textit{All}} & \textbf{ \textit{Types}} & \textbf{ \textit{Arguments}}  & \textbf{\popper}\\
\midrule
\emph{md} & \textbf{100 $\pm$ 0} & \textbf{100 $\pm$ 0} & \textbf{100 $\pm$ 0} & \textbf{100 $\pm$ 0}\\
\emph{buttons-next} & 98 $\pm$ 0 & \textbf{100 $\pm$ 0} & \textbf{100 $\pm$ 0} & \textbf{100 $\pm$ 0}\\
\emph{coins-next} & 92 $\pm$ 1 & \textbf{100 $\pm$ 0} & \textbf{100 $\pm$ 0} & \textbf{100 $\pm$ 0}\\
\emph{buttons-goal} & \textbf{100 $\pm$ 0} & \textbf{100 $\pm$ 0} & \textbf{100 $\pm$ 0} & 96 $\pm$ 1\\
\emph{coins-goal} & 96 $\pm$ 0 & \textbf{100 $\pm$ 0} & \textbf{100 $\pm$ 0} & \textbf{100 $\pm$ 0}\\
\emph{gt-centipede-goal} & \textbf{82 $\pm$ 0}& 75 $\pm$ 0 & 75 $\pm$ 0 & 75 $\pm$ 0 \\
\emph{gt-centipede-legal} & \textbf{100 $\pm$ 0}&\textbf{100 $\pm$ 0} &\textbf{100 $\pm$ 0} & \textbf{100 $\pm$ 0} \\
\emph{gt-centipede-next} & \textbf{100 $\pm$ 0}& \textbf{100 $\pm$ 0}& \textbf{100 $\pm$ 0}& \textbf{100 $\pm$ 0}\\
\emph{krk} & \textbf{98 $\pm$ 0} & \textbf{98 $\pm$ 0} & \textbf{98 $\pm$ 0} & \textbf{98 $\pm$ 0}\\
\emph{list} & 50 $\pm$ 0 & \textbf{100 $\pm$ 0} & \textbf{100 $\pm$ 0} & 50 $\pm$ 0 \\
\emph{powerof2} & \textbf{100 $\pm$ 0} & \textbf{100 $\pm$ 0} & \textbf{100 $\pm$ 0} & 84 $\pm$ 1 \\
\emph{append} & 95 $\pm$ 1 & 95 $\pm$ 1 & \textbf{96 $\pm$ 1} & \textbf{96 $\pm$ 1}\\

\end{tabular}
\caption{
Predictive Accuracy. We round to the closest integer. The error is standard deviation.}
\label{tab:acc_bias}
\end{table}
\subsubsection{Results}
Table \ref{tab:time_bias} shows the learning times.
These results show that in general, \textit{All} requires learning times longer than or equal to \textit{Types}, which in turn requires learning times longer than or equal to \textit{Arguments}. For instance, \textit{All} reaches timeout on the task \textit{list}, while \textit{Types} and \textit{Arguments} require respectively 2s and 1s. In some experiments such as \emph{krk}, \textit{Types} and \textit{Arguments} have equivalent bias, because the arguments specified are the only arguments of the types specified. 
\popper{} is provided with types for the unary predicates in these domains and thus rather is comparable with \textit{Types}. Yet, \textit{Types} outperforms \popper{} in terms of learning times. This result can be explained by the fact that \textit{Types} is a variant of \name. As such, it considers magic hypotheses which represent the set of their instantiations. Therefore, in contrast to \popper{}, \textit{Types} does not enumerate all possible candidate constants. Moreover, \textit{Types} only considers instantiations which, together with the background knowledge, entail at least one positive example, while \popper{} considers every possible constant in the search space equally. Also, because \textit{Types} does not require additional unary predicates, it can express hypotheses more compactly and can search up to a smaller depth. \popper{} can also achieve longer learning times than \textit{All} whereas \textit{All} searches a larger space. For instance, \popper{} requires 18s to solve the task \emph{powerof2} while \textit{All} solves it in less than 1s.

Table \ref{tab:acc_bias} shows the predictive accuracies.
These results show \textit{All} can achieve lower predictive accuracies than \textit{Types} and \textit{Arguments}. For instance, \textit{All} reaches 92\% accuracy on \textit{coins-next} while \textit{Types} and \textit{Arguments} reach 100\% accuracy. There are two main reasons explaining this difference. First, \textit{All} has a more expressive language, and in particular can express more specific hypotheses through the use of more constant symbols. It is thus more prone to overfitting. Second, \textit{All} searches a larger search space. It consequently might not find an optimal solution before timeout. Moreover, according to the Blumer bound \cite{blumer1989}, searching a larger search space can result in lower predictive accuracies.

We can conclude that \name{} can benefit from additional bias about which variables should be magic variables, and in particular it can achieve better learning performance. We thus can positively answer \textbf{Q4}. This experiment illustrates the impact of more bias. More bias can help reduce the search space and thus improve learning performance. However, this bias must be user provided.
More generally, choosing an appropriate bias is a key challenge in ILP \cite{cropper2020c}.






\subsection[\textbf{Q5}: unnecessary magic values]{\textbf{Q5}: effect on learning performance for problems which do not require magic values}\label{q5}
Our approach can improve learning performance for problems which require magic values. 
However, magic values are not always required and it is not always known whether a good solution requires magic values. 
We investigate in this experiment the impact on learning performance of unnecessarily allowing magic values and thus aim to answer \textbf{Q5}.

\subsubsection{Material and Methods}
To answer \textbf{Q5}, we compare systems which allow constant symbols with systems which disallow constant symbols. Since it is unknown which variables should be constant symbols, we allow any variable to be a constant symbol. Therefore, we use the \textit{All} setting for \name. We call \ale$_{c}$ the version of \ale{} for which any argument is allowed to be a constant symbol, and \ale$_{\cancel{c}}$ the version of \ale{} which disallows constant symbols. At the end, we compare \name{} with \popper{} and \ale$_{c}$ with \ale$_{\cancel{c}}$. 
We use a timeout of 600s per task. We consider two different domains.
\paragraph{Michalski trains.} The goal of these tasks is to find a hypothesis that distinguishes eastbound and westbound trains \cite{larson1977}. We use four increasingly complex tasks. There are 1000 examples but the distribution of positive and negative examples is different for each task. We randomly sample the examples and split them into 80/20 train/test partitions.
\paragraph{Program synthesis: \textit{evens}, \textit{last}, \textit{member}, \textit{sorted}.} We use the same material and methods as \cite{cropper2020b}. These problems all involve learning recursive hypotheses.
\subsubsection{Results}
Tables \ref{tab:time_no_magic_values} and \ref{tab:acc_no_magic_values} show the results. They show \popper{} always outperforms \name{} and \ale$_{\cancel{c}}$ always outperforms \ale$_{c}$ in terms of learning times. For instance, \name{} takes 7s when solving the \textit{trains1} task while \popper{} solves it in 3s. \ale$_{c}$ reaches timeout on all \emph{trains} tasks while \ale$_{\cancel{c}}$ solves them in a few seconds. Because it searches a larger space, \name{} and \ale$_{c}$ require longer learning times compared to \popper{} and \ale$_{\cancel{c}}$ respectively.

This increase in learning time can reduce predictive accuracies since \name{} or \ale$_{c}$ consequently might not find a solution before timeout. For instance, \ale$_{c}$ reaches timeout and thus achieves the default accuracy on the \textit{trains} tasks while \ale{} achieves maximal accuracy. Moreover, \name{} and \ale$_{c}$ are more prone to overfitting. For instance, \name{} learns overly specific hypotheses for \textit{last}. 
Finally, according to the Blumer bound \cite{blumer1989}, searching a larger space can result in lower predictive accuracies. 
However, allowing constant symbols in programs provides better expressivity, since more hypotheses can be formulated compared to disallowing constant symbols. In particular, it might allow the learner to formulate more accurate hypotheses.
For instance, \ale$_{c}$ achieves better accuracies than \ale$_{\cancel{c}}$ on \emph{member} and \emph{sorted}.

To conclude, these results show that allowing magic values when unnecessary can impair learning performance, in particular learning times and predictive accuracies, which answers \textbf{Q5}. Future work is needed to automatically identify when magic values are necessary and which variables could be magic values. More generally, identifying a suitable bias is a major challenge in ILP \cite{cropper2020c}.

\begin{table}[ht]
\centering
\begin{tabular}{l|cccc}
\textbf{Task} & \textbf{\ale$_{c}$} & \textbf{\ale$_{\cancel{c}}$} & \textbf{\name} & \textbf{\popper}\\
\midrule
\emph{trains1} & timeout & \textbf{1 $\pm$ 0} & 7 $\pm$ 0 & 3 $\pm$ 0\\
\emph{trains2} & timeout & \textbf{1 $\pm$ 0} & 4 $\pm$ 0 & 3 $\pm$ 0\\
\emph{trains3} & timeout & \textbf{2 $\pm$ 0} & 31 $\pm$ 0 & 24 $\pm$ 0\\
\emph{trains4} & timeout & \textbf{5 $\pm$ 0} & 25 $\pm$ 0 & 21 $\pm$ 0\\
\emph{evens}  & 14 $\pm$ 1 & \textbf{0 $\pm$ 0} & 5 $\pm$ 1 & 1 $\pm$ 0\\
\emph{last}  & 16 $\pm$ 1 & \textbf{0 $\pm$ 0} & 180 $\pm$ 91 & 1 $\pm$ 0\\
\emph{member} & 15 $\pm$ 1 & \textbf{0 $\pm$ 0} & \textbf{0 $\pm$ 0} &\textbf{0 $\pm$ 0}\\
\emph{sorted} & 5 $\pm$ 1 & \textbf{0 $\pm$ 0} &  86 $\pm$ 24 & 70 $\pm$ 58\\
\end{tabular}
\caption{
Learning times. We round times to the nearest second. The error is standard deviation.}
\label{tab:time_no_magic_values}
\end{table}

\begin{table}[ht]
\centering
\begin{tabular}{l|cccc}
\textbf{Task} & \textbf{\ale$_{c}$} & \textbf{\ale$_{\cancel{c}}$} & \textbf{\name} & \textbf{\popper}\\
\midrule
\emph{trains1} & 50 $\pm$ 0 & \textbf{100 $\pm$ 0} & \textbf{100 $\pm$ 0} &\textbf{100 $\pm$ 0}\\
\emph{trains2} & 50 $\pm$ 0 & 98 $\pm$ 2 & \textbf{99 $\pm$ 0} & \textbf{99 $\pm$ 0}\\
\emph{trains3} & 50 $\pm$ 0 & \textbf{100 $\pm$ 0} & \textbf{100 $\pm$ 0} &\textbf{100 $\pm$ 0}\\
\emph{trains4} & 50 $\pm$ 0 & \textbf{100 $\pm$ 0} & \textbf{100 $\pm$ 0} &\textbf{100 $\pm$ 0}\\
\emph{evens}  & 51 $\pm$ 0 & 54 $\pm$ 4 & \textbf{100 $\pm$ 0} &\textbf{100 $\pm$ 0}\\
\emph{last}  & 50 $\pm$ 0 & 50 $\pm$ 0 & 85 $\pm$ 7 &\textbf{100 $\pm$ 0}\\
\emph{member} & 53 $\pm$ 1 & 50 $\pm$ 0 & \textbf{100 $\pm$ 0} &\textbf{100 $\pm$ 0}\\
\emph{sorted} & 75 $\pm$ 2 & 72 $\pm$ 3 & 90 $\pm$ 5 &\textbf{97 $\pm$ 2}\\
\end{tabular}
\caption{
Predictive Accuracy. We round to the closest integer. The error is standard deviation.}
\label{tab:acc_no_magic_values}
\end{table}

\section{Conclusion and Limitations}


 Learning programs with magic values is fundamental to many AI applications. 
 However, current program synthesis approaches rely on enumerating candidate constant symbols, which inhibits scalability and prohibits learning them from continuous domains. 
 To overcome this limitation, we have introduced an ILP approach to efficiently learn programs with magic values from potentially large or infinite domains. 
 Inspired by \ale's lazy evaluation procedure \cite{srinivasan1999}, our approach builds partial hypotheses with variables in place of constant symbols. 
 Therefore, our approach does not enumerate all candidate constants when constructing hypotheses. 
 The particular constant symbols are identified by executing the hypothesis on the examples. Thus, our approach only considers relevant constant symbols which can be obtained from the examples. Our approach extends the LFF framework with constraints to prune partial hypotheses, which each represent a set of instantiated hypotheses. 
 For these reasons, our approach can efficiently learn in large, potentially infinite, domains. 
Our experiments on several domains show that our approach can (i) outperform state-of-the-art approaches and (ii) scale to domains with millions of constant symbols and even infinite ones, including continuous domains. 


\subsection*{Limitations and Future Work}

\paragraph{Noise.} 
In contrast to other ILP systems \cite{karalivc1997,blockeel1998,srinivasan2001}, \name{} cannot identify magic values from noisy examples. Previous work \cite{wahlig2022} has extended LFF to support learning from noisy examples by relaxing the completeness and consistency conditions as well as hypotheses constraints applications. This extension should be directly applicable to \name, which we will address as future work.

\paragraph{Scalability.}


To find magic values in a same clause, our approach must search through the cartesian product of each potential magic value. 
Therefore, its scalability is limited when increasing the number of magic values in the same clause, as shown in the experiment presented in Section \ref{oneclause}. Also, \name{} finds candidate constant symbols from executing hypotheses on the training examples. As shown in Section \ref{examples}, its scalability is limited when increasing the number of examples.

\paragraph{Bias.}
\name{} can be provided as bias which variables can be bound to constant symbols, if this bias is known, or can automatically identify these variables at the expense of more search. Our experiments presented in Sections \ref{bias} and \ref{q5} have shown that without this additional bias, learning performance can be degraded, in particular learning times and predictive accuracies. Choosing an appropriate bias more generally is a major issue with ILP systems \cite{cropper2020c}. As far as we are aware, no system can automatically identify suitable bias which future work should address.

\paragraph{Numerical values.}
Our magic value evaluation procedure identifies bindings by executing the hypothesis over each example independently. Therefore, it can only find magic values which value arises from single positive examples. In particular, it cannot identify magic values for which multiple examples are required for their evaluation. For example, it cannot identify parameters of linear or polynomial equations in contrast to other ILP systems \cite{karalivc1997,srinivasan1999}. 
Likewise, it cannot identify values requiring numerical reasoning, such as identifying an optimal threshold \cite{blockeel1997,srinivasan1999}. 
For the same reason, our method cannot create new constant symbols which are not part of the domain.
To overcome this limitation, we plan to use SMT solvers to identify magic values from reasoning from multiple examples, positive and negative. 





\section*{Declarations}
This research was funded in whole, or in part, by the EPSRC grant \emph{Explainable Drug Design} and the EPSRC fellowship \emph{The Automatic Computer Scientist} (EP/V040340/1). For Open Access, the author has applied a CC BY public copyright licence to any Author Accepted Manuscript version arising from this submission. All data supporting this study is provided as supplementary information accompanying this paper.

\section*{Acknowledgments}
The authors thank H\r{a}kan Kjellerstrand, Rolf Morel, and Oghenejokpeme Orhobor for valuable feedback.

\bibliography{bibliography}
\begin{appendices}

\section{Proofs}
\subsection{Extended Constraints}
\setcounter{lemma}{0}
\setcounter{proposition}{0}We first state three lemmas. These lemmas justify why we can restrict the search for instantiations to instantiations which cover at least one positive example.

\begin{lemma}
Let $(E^{+},E^{-},B,{\cal{H}},C)$ be an LFF input and $H$ a magic hypothesis, if all the instantiations $I$ of $H$ such that $\exists e \in E^{+}, B\cup I \models e$ are incomplete, then all instantiations of $H$ are incomplete. \label{lemma2}
\end{lemma}
\begin{proof1}
Instantiations $I$ such that $\not\exists e \in E^{+}, B\cup I \models e$ are incomplete.
\end{proof1}
\begin{lemma}
Let $(E^{+},E^{-},B,{\cal{H}},C)$ be an LFF input and $H$ a magic hypothesis, if all the instantiations $I$ of $H$ such that $\exists e \in E^{+}, B\cup I \models e$ are inconsistent, then all instantiations of $H$ are totally incomplete or inconsistent. \label{lemma3}
\end{lemma}
\begin{proof1}
Instantiations $I$ such that $\not\exists e \in E^{+}, B\cup I \models e$ are totally incomplete.
\end{proof1}
\begin{lemma}
Let $(E^{+},E^{-},B,{\cal{H}},C)$ be an LFF input and $H$ a magic hypothesis, if $H$ has no instantiation $I$ such that $\exists e \in E^{+}, B\cup I \models e$, then all instantiations of $H$ are totally incomplete.\label{lemma1}
\end{lemma}
\begin{proof1}
Instantiations $I$ such that $\not\exists e \in E^{+}, B\cup I \models e$ are totally incomplete.
\end{proof1}
\noindent
We now introduce extensions of specialisation, generalisation, and redundancy constraints which prune magic hypotheses. We use the three lemmas above to prove these constraints are optimally sound.



\begin{proposition}
Let $(E^{+},E^{-},B,{\cal{H}},C)$ be an LFF input, $H_1 \in {\cal{H}}_C$, and $H_2 \in {\cal{H}}_C$ be two magic hypotheses such that $H_1 \preceq H_2$. 
If all instantiation $I_1$ of $H_1$ such that $\exists e \in E^{+}, B\cup I_1 \models e$ are incomplete, then all instantiation of $H_2$ also are incomplete.
\end{proposition}
\begin{proof1}
All instantiations $I_1$ of $H_1$ such that $\exists e \in E^{+}, B\cup I_1 \models e$ are incomplete. Therefore, according to Lemma \ref{lemma2}, all instantiations of $H_1$ are incomplete. Let $H_2$ be a specialisation of $H_1$. Let $I_2$ be an instantiation of $H_2$. $I_2$ is a specialisation of an instantiation $I_1$ of $H_1$. $I_1$ is incomplete. Since subsumption implies entailment, then $I_2$ also is incomplete.
\end{proof1}

\begin{proposition}
Let $(E^{+},E^{-},B,{\cal{H}},C)$ be an LFF input, $H_1 \in {\cal{H}}_C$ and $H_2 \in {\cal{H}}_C$ be two magic hypotheses such that $H_2$ is non-recursive and $H_2 \preceq H_1$. If all instantiation $I_1$ of $H_1$ such that $\exists e \in E^{+}, B\cup I_1 \models e$ are inconsistent, then all instantiations of $H_2$ are inconsistent or non-optimal.
\end{proposition}
\begin{proof1}
Let $H_2$ be a non-recursive generalisation of $H_1$. Let $I_2$ be an instantiation of $H_2$. $I_2$ is a generalisation of an instantiation $I_1$ of $H_1$. By assumption, all instantiations $I_1$ of $H_1$ such that $\exists e \in E^{+}, B\cup I_1 \models e$ are inconsistent. Therefore, according to Lemma \ref{lemma3}, all instantiations of $H_1$ are (1) inconsistent or (2) totally incomplete. For (1), if $I_1$ is inconsistent, then $I_2$ also is inconsistent. For (2), assume $I_1$ is totally incomplete. Since $I_2$ is a generalisation of $I_1$ and is non-recursive, then $I_1$ is an independent subset of $I_2$ of totally incomplete clauses. Therefore, $I_1$ is a subset of clauses which is redundant in $I_2$ and thus $I_2$ is non-optimal.
\end{proof1}

\begin{proposition}
Let $(E^{+},E^{-},B,{\cal{H}},C)$ be an LFF input, $H_1 \in {\cal{H}}_C$ be a magic hypothesis. If $H_1$ has no instantiation $I_1$ such that $\exists e \in E^{+}, B\cup I_1 \models e$, then all non-recursive magic hypotheses $H_2$ which contain a specialisation of $H_1$ as a subset are non-optimal.
\end{proposition}
\begin{proof1}
Let $H_2$ be a non-recursive hypothesis which contains a specialisation of $H_1$ as a subset. Let $I_2$ be an instantiation of $H_2$. Then $I_2$ contains as a subset a specialisation of an instantiation $I_1$ of $H_1$. $H_1$ has no instantiation $I_1$ such that $\exists e \in E^{+}, B\cup I_1 \models e$, therefore, according to Lemma \ref{lemma1}, all instantiations of $H_1$ are totally incomplete. Therefore, $I_1$ is totally incomplete. Moreover, $I_2$ is non-recursive by assumption. Then $I_1$ is an independent subset of redundant clauses in $I_2$ and thus $I_2$ is non-optimal.
\end{proof1}
\subsection{Theoretical Analysis}
The following proposition evaluates the reduction over the hypothesis space of using magic values instead of enumerating all candidate constant symbols as unary predicates.
\begin{proposition}
Let $D_b$ be the number of body predicates available in the search space, $m$ be the maximum number of body literals allowed in a clause, $c$ the number of constant symbols available, and $n$ the maximum number of clauses allowed in a hypothesis. Then the maximum number of hypotheses in the hypothesis space can be multiplied by a factor of $(\frac{D_b+c}{D_b})^{mn}$ if representing constants with unary predicate symbols, one per allowed constant symbol, compared to using magic variables. \label{prop:gain}
\end{proposition}

\begin{proof1}
Let Let $D_h$ be the number of head predicates in the search space, $a$ be the maximum arity of predicates and $v$ be the maximum number of unique variables allowed in a clause. From \cite{cropper2020b}, the maximum number of hypotheses in the hypothesis space is:
\begin{align*}
\sum_{j=1}^{n}\binom{\mid D_h \mid v^a \sum_{i=1}^{m} \binom{\mid D_b \mid v^a}{i}}{j} 
\end{align*}
Given that $0 \leq i < \mid D_b \mid v^a$, we have:
\begin{align*}
\binom{\mid D_b \mid v^a}{i} \leq \frac{(\mid D_b \mid v^a)^i}{i!}
\end{align*}
Therefore:
\begin{align*}
\mid D_h \mid v^a \sum_{i=1}^{m} \binom{\mid D_b \mid v^a}{i} < \mid D_h \mid v^a m (\mid D_b \mid v^a)^m
\end{align*}
And similarly, we have: 
\begin{align*}
\sum_{j=1}^{n}\binom{\mid D_h \mid v^a \sum_{i=1}^{m} \binom{\mid D_b \mid v^a}{i}}{j} \leq n(\mid D_h \mid v^a m (\mid D_b \mid v^a)^m)^n
\end{align*}
If adding one unary predicate symbol per constant symbol, then there are $D_b+c$ body predicates available. Then, the maximum number of hypotheses in the hypothesis space above becomes:
\begin{align*}
\sum_{j=1}^{n}\binom{\mid D_h \mid v^a \sum_{i=1}^{m} \binom{\mid (D_b+c) \mid v^a}{i}}{j} \leq n(\mid D_h \mid v^a m (\mid (D_b+c) \mid v^a)^m)^n
\end{align*}
Therefore, representing constants with magic variables can reduce the maximum size of the hypothesis space by a factor of:
\begin{align*}
(\frac{D_b+c}{D_b})^{mn}
\end{align*}
\end{proof1}

\section{Experiments}
\subsection{Domains}
We describe the domains used in our experiments.
\paragraph{IGGP} Figures \ref{fig:coins}, \ref{fig:coins-goal}, \ref{fig:buttons_next}, \ref{fig:gt-centipede-goal}, \ref{fig:gt-centipede-legal} and \ref{fig:gt-centipede-next} represent some example solutions for these tasks.

\begin{figure}[t]
\begin{tabular}{l}
\emph{next\_cell(A,B,C) $\leftarrow$ my\_true\_cell(A,B,C),does\_jump(A,\textbf{robot},F,D),}\\
\emph{different(B,F),different(D,B)}\\
\emph{next\_cell(A,B,\textbf{twocoins})$\leftarrow$ does\_jump(A,E,F,D),different(D,F),}\\
\emph{does\_jump(A,E,F,B)}\\
\emph{next\_cell(A,B,\textbf{zerocoins}) $\leftarrow$ does\_jump(A,F,B,E),does\_jump(A,F,D,E),}\\
\emph{different(E,D)}
\end{tabular}
\caption{
 Example solution for the IGGP \emph{coins-next} task. Magic values are represented in bold.}
\label{fig:coins}
\end{figure}

\begin{figure}[t]
\begin{tabular}{l}
\emph{goal(A,\textbf{robot},\textbf{100})$\leftarrow$ my\_true\_step(A,\textbf{5})}\\
\emph{goal(A,\textbf{robot},\textbf{0})$\leftarrow$ my\_true\_cell(A,D,\textbf{onecoin}),my\_true\_cell(A,D,\textbf{onecoin})}
\end{tabular}
\caption{
 Example solution for the IGGP \emph{coins-goal} task. Magic values are represented in bold.}
\label{fig:coins-goal}
\end{figure}

\begin{figure}[t]
\begin{tabular}{l}
\emph{goal(A,\textbf{robot},\textbf{100}) $\leftarrow$ my\_true(A,\textbf{p}),my\_true(A,\textbf{r}),my\_true(A,\textbf{q})}\\
\emph{goal(A,\textbf{robot},\textbf{0}) $\leftarrow$ not\_my\_true(A,\textbf{r})}\\
\emph{goal(A,\textbf{robot},\textbf{0}) $\leftarrow$ not\_my\_true(A,\textbf{q})}\\
\emph{goal(A,\textbf{robot},\textbf{0}) $\leftarrow$ not\_my\_true(A,\textbf{p})}\\
\end{tabular}
\caption{
 Example solution for the \textit{buttons-goal} task. Magic values are represented in bold.}
\label{fig:buttons_next}
\end{figure}

\begin{figure}[t]
\begin{tabular}{l}
\emph{goal(A,B,\textbf{0}) $\leftarrow$ true\_blackPayoff(A,\textbf{25}),role(B).}\\
\emph{goal(A,B,\textbf{0}) $\leftarrow$ true\_blackPayoff(A,\textbf{35}),role(B).}\\
\emph{goal(A,B,\textbf{0}) $\leftarrow$ true\_blackPayoff(A,\textbf{45}),role(B).}\\
\emph{goal(A,B,\textbf{0}) $\leftarrow$ true\_whitePayoff(A,\textbf{0}),role(B),true\_control(A,\textbf{black})}\\
\emph{goal(A,\textbf{white},\textbf{0}) $\leftarrow$ true\_control(A,\textbf{white})}\\
\emph{goal(A,\textbf{black},\textbf{0}) $\leftarrow$ true\_whitePayoff(A,\textbf{35})}\\
\emph{goal(A,\textbf{black},\textbf{0}) $\leftarrow$ true\_control(A,\textbf{white}),true\_whitePayoff(A,\textbf{15})}\\
\emph{goal(A,\textbf{black},\textbf{15}) $\leftarrow$ true\_whitePayoff(A,\textbf{0}),true\_control(A,\textbf{white})}\\
\emph{goal(A,\textbf{black},\textbf{0}) $\leftarrow$ true\_whitePayoff(A,\textbf{5})}\\
\emph{goal(A,\textbf{black},\textbf{10}) $\leftarrow$ true\_control(A,\textbf{black}),true\_whitePayoff(A,\textbf{15})}\\
\emph{goal(A,\textbf{white},C) $\leftarrow$ true\_control(A,\textbf{black}),true\_blackPayoff(A,D),}\\
\hspace{89pt}\emph{true\_whitePayoff(A,C),my\_succ(D,C)}\\
\end{tabular}
\caption{
 Example solution for the \textit{gt-centipede-goal} task. Magic values are represented in bold.}
\label{fig:gt-centipede-goal}
\end{figure}

\begin{figure}[t]
\begin{tabular}{l}
\emph{legal(A,B,\textbf{finish}) $\leftarrow$ true\_control(A,B)}\\
\emph{legal(A,\textbf{black},\textbf{noop}) $\leftarrow$ true\_control(A,\textbf{white})}\\
\emph{legal(A,\textbf{white},\textbf{noop}) $\leftarrow$ true\_control(A,\textbf{black})}\\
\emph{legal(A,B,\textbf{continue}) $\leftarrow$ true\_control(A,B)}\\
\end{tabular}
\caption{
 Example solution for the \textit{gt-centipede-legal} task. Magic values are represented in bold.}
\label{fig:gt-centipede-legal}
\end{figure}

\begin{figure}[t]
\begin{tabular}{l}
\emph{next\_control(A,\textbf{white}) $\leftarrow$ true\_control(A,\textbf{black})}\\
\emph{next\_control(A,\textbf{black}) $\leftarrow$ true\_control(A,\textbf{white})}\\
\end{tabular}
\caption{
 Example solution for the \textit{gt-centipede-next} task. Magic values are represented in bold.}
\label{fig:gt-centipede-next}
\end{figure}

\paragraph{KRK} We provide each system with the quadratic predicate \textit{cell} and the triadic predicate \textit{distance}. For \name, any variable of type \textit{type}, \textit{color} or \textit{int} is allowed to be a magic value. Training sets contain 10 positive examples and 10 negative ones.

\paragraph{Program synthesis: \emph{list}}
 Positive examples are generated by randomly choosing a position for the element $7$ and sampling the remaining elements from the constants available in the background knowledge. Negative examples are generated by randomly sampling elements from the constants in the background knowledge. Lists have size 500. We generate 10 positive and 10 negative training examples. 
Therefore, the default accuracy is 50\%. We provide each learning system with the dyadic predicates \textit{head}, \textit{tail}, \textit{length}, \textit{last}, \textit{geq}, and the monadic predicate \textit{empty}. For \name, any variable of type \textit{element} is allowed to be a magic value. 

\begin{figure}[t]
\begin{tabular}{l}
\emph{multiple(\textbf{1})}\\
\emph{multiple(A) $\leftarrow$ div(A,\textbf{2},B),multiple(B)}\\
\end{tabular}
\caption{
 Example solution for the \textit{powerof2} task. Magic values are represented in bold.}
\label{fig:powerof2}
\end{figure}

\paragraph{Program synthesis: \emph{powerof2}} Figure \ref{fig:powerof2} represents an example of target hypothesis. Positive examples are powers of 2 between 2 and $2^{10}$. Negative examples are numbers between 2 and $2^{10}$ which are not a power of 2. We sample 10 positive and 10 negative examples. We provide each system with the triadic predicate \textit{div}. For \name, any variable is of type \textit{number} and thus can be a magic variable.

\paragraph{Program synthesis: \emph{append}} Examples are lists of size 10 of elements from the set of available constants. A magic suffix of size 2 is randomly sampled from the set of constants. Positive examples are lists ending with the magic suffix. Negative examples are lists that do not end with the magic suffix. We sample 10 positive and 10 negative examples. We provide each system with the triadic predicate \textit{append}, the dyadic predicates \textit{head} and \textit{tail}. For \name, any variable is of type list and thus can be a magic variable.

\paragraph{Learning Pi}
An example is a pair radius / area. Radius are real numbers sampled between 0 and 10. We sample 20 training examples, half positive and half negative. We provide each system with the triadic predicates \textit{add}, \textit{subtract}, \textit{multiply}, \textit{divide} and the dyadic predicate \textit{square}. 

\paragraph{Equilibrium}
An example is an object, its mass and the forces it is subject to are real numbers sampled between 0 and 10. There are 7 forces applied to an object. 
We sample 10 training examples, half positive and half negative. We provide each system with the triadic predicates \textit{add}, \textit{subtract}, \textit{multiply}, \textit{divide} and the dyadic predicates \textit{sum}, \textit{square}, \textit{mass} and \textit{force}. For \name, any variable of type \textit{number} is allowed to be a magic value.

\paragraph{Drug design}
Each molecule contains 10 atoms, pairwise distances are floats sampled between 0 and 10. There are 10 different atom types in the domain. A positive example contains two atoms, one of type \textit{o} and one of type \textit{h}, separated by the target distance. Target distances are randomly generated for each run. We sample 20 training examples, half positive and half negative. We provide each system with the triadic predicate \textit{distance} and the dyadic predicates \textit{atom} and \textit{atom$\_$type}. For \name, any variable of type \textit{type} or \textit{number} is allowed to be a magic value.

\paragraph{Program Synthesis: \emph{next}}
 The target magic value and list elements are real numbers chosen at random. An example is a pair list/value where the value is an element of the list. We provide each learning system with the dyadic background predicates \textit{head}, \textit{tail}, \textit{length}, \textit{last}, \textit{geq} and the monadic predicate \textit{empty}.  For \name, any variable of type \textit{number} is allowed to be a magic value. We sample 20 training examples, half positive and half negative.

\paragraph{Program Synthesis: \emph{sumk}}
 The target sum and list elements are integers chosen at random. We provide each learning system with the dyadic background predicates \textit{member} and triadic predicate \textit{sum}. For \name, any variable of type \textit{number} is allowed to be a magic value. We sample 20 training examples, half positive and half negative.

\subsection{Material and methods}
We describe our experimental designs.
\paragraph{Minimal Decay}
We generate an ordered set of constant symbols $S$ of varying size. An example is a state described by a true value from $S$, a next value from $S$ and a player action from $\{$\emph{press$\_$button}, \emph{noop}$\}$. True values and player actions are chosen at random. For each positive example, a set of negative examples are generated by taking a subset of size at most 1000 of the available constants as next value for which the target theory does not hold. We generate 10 positive training examples for each task.

\paragraph{Magic values in one clause}
 We generate 10 positive and 10 negative examples. For successive values of $n$, we sample a set of $n$ magic values. To generate a positive example, we assign $n$ different position in the list, one for each of the magic values. Other values in the list are chosen at random among the set of other available constant symbols. To generate a negative example, we choose $n-1$ magic values, which we assign to $n-1$ different position in the list. Other values are chosen at random among the set of other available constant symbols. We allow the dyadic body predicate \textit{member} in the background knowledge.

\paragraph{Magic values in multiple clauses}
 We sample a set $S$ of constant symbols, among which we sample a subset $M$ of varying size representing target magic values. Examples are lists of elements from $S$. A positive example is an example which head element is in $M$. We generate 200 positive and 200 negative training examples. We check that each training positive example set contains at least one example for each clause to ensure the target is identifiable.
 
 \paragraph{Program synthesis} We give each system the following dyadic relations \textit{head}, \textit{tail}, \textit{decrement}, \textit{geq} and the monadic relations \textit{empty}, \textit{zero}, \textit{one}, \textit{even}, and \textit{odd}.
\end{appendices}

\end{document}